\documentclass[twocolumn, switch]{article} 

\usepackage{comment}
\usepackage{svg}
\usepackage{bbding}
\usepackage{preprint}
\usepackage{pifont}
\usepackage{xcolor}
\usepackage{gensymb}

\usepackage{siunitx}

\usepackage{xspace}
\makeatletter
\DeclareRobustCommand\onedot{\futurelet\@let@token\@onedot}

\def\@onedot{\ifx\@let@token.\else.\null\fi\xspace}

\def\eg{\emph{e.g}\onedot} 
\def\ie{\emph{i.e}\onedot}

\def\etal{\emph{et al}\onedot}
\makeatother

\usepackage{amsmath, amsthm, amssymb, amsfonts}

\usepackage[numbers,square]{natbib}
\usepackage[utf8]{inputenc}	
\usepackage[T1]{fontenc}	
\usepackage{xcolor}		
\usepackage[colorlinks = true,
            linkcolor = purple,
            urlcolor  = blue,
            citecolor = cyan,
            anchorcolor = black]{hyperref}	
\usepackage{booktabs} 		
\usepackage{nicefrac}		
\usepackage{microtype}		
\usepackage{lineno}		
\usepackage{float}			

\usepackage{lipsum}		

\usepackage{newfloat}
\DeclareFloatingEnvironment[name={Supplementary Figure}]{suppfigure}
\usepackage{sidecap}
\sidecaptionvpos{figure}{c}

\usepackage{titlesec}
\titlespacing\section{0pt}{12pt plus 3pt minus 3pt}{1pt plus 1pt minus 1pt}
\titlespacing\subsection{0pt}{10pt plus 3pt minus 3pt}{1pt plus 1pt minus 1pt}
\titlespacing\subsubsection{0pt}{8pt plus 3pt minus 3pt}{1pt plus 1pt minus 1pt}

\usepackage{tikz,xcolor,hyperref}

\usepackage{cleveref}

\usepackage{dblfloatfix}

\definecolor{lime}{HTML}{A6CE39}
\DeclareRobustCommand{\orcidicon}{
	\begin{tikzpicture}
	\draw[lime, fill=lime] (0,0) 
	circle [radius=0.16] 
	node[white] {{\fontfamily{qag}\selectfont \tiny ID}};
	\draw[white, fill=white] (-0.0625,0.095) 
	circle [radius=0.007];
	\end{tikzpicture}
	\hspace{-2mm}
}
\foreach \x in {A, ..., Z}{\expandafter\xdef\csname orcid\x\endcsname{\noexpand\href{https://orcid.org/\csname orcidauthor\x\endcsname}
			{\noexpand\orcidicon}}
}

\title{Domain Generalization for In-Orbit 6D Pose Estimation}

\usepackage{authblk}
\author[1,2,4\thanks{\tt{antoine.legrand@uclouvain.be}}]{Antoine Legrand\orcidA{}}
\author[2,3]{Renaud Detry\orcidB{}}
\author[1]{Christophe De Vleeschouwer \orcidC{}}

\affil[1]{UCLouvain, ICTEAM, Dept. Electrical Engineering}
\affil[2]{KU Leuven, Dept. Electrical Engineering}
\affil[3]{KU Leuven, Dept. Mechanical Engineering}
\affil[4]{Aerospacelab}

\begin{document}

\twocolumn[ 
  \begin{@twocolumnfalse} 
  
\maketitle

\begin{abstract}
We address the problem of estimating the relative 6D pose, \ie, position and orientation, of a target spacecraft, from a monocular image, a key capability for future autonomous Rendezvous and Proximity Operations. Due to the difficulty of acquiring large sets of real images, spacecraft pose estimation networks are exclusively trained on synthetic ones. However, because those images do not capture the illumination conditions encountered in orbit, pose estimation networks face a domain gap problem, \ie, they do not generalize to real images. Our work introduces a method that bridges this domain gap. It relies on a novel, end-to-end, neural-based architecture as well as a novel learning strategy. This strategy improves the domain generalization abilities of the network through multi-task learning and aggressive data augmentation policies, thereby enforcing the network to learn domain-invariant features. We demonstrate that our method effectively closes the domain gap, achieving state-of-the-art accuracy on the widespread SPEED+ dataset. Finally, ablation studies assess the impact of key components of our method on its generalization abilities.
\end{abstract}
\vspace{0.35cm}

  \end{@twocolumnfalse} 
] 

\section{Introduction}   
    Future space missions such as space debris mitigation~\cite{forshaw2016removedebris,reed2016restore,biesbroek2021clearspace} or on-orbit servicing~\cite{henshaw2014darpa,pyrak2022mevs} (\eg, refueling, repairing or inspection of end-of-life satellites) are gaining in interest among both space agencies~\cite{henshaw2014darpa,forshaw2016removedebris,reed2016restore} and private companies~\cite{biesbroek2021clearspace,pyrak2022mevs}. These missions involve Rendezvous and Proximity Operations (RPOs) between two spacecraft. While some previous missions conducted RPOs via tele-operation, autonomous operations are preferred nowadays because they are considered safer and cheaper. This motivates the development of Guidance Navigation \& Control (GNC) systems that allow a spacecraft to navigate by itself at close range of its target~\cite{ventura2016autonomous}. A key component of this GNC system is the Navigation subsystem that has to estimate the 6D pose, \ie, position and orientation, of the target spacecraft relative to the servicer.
    
    The complexity of this estimation depends on whether the RPO is cooperative or uncooperative. In the former case, the target spacecraft is equipped with known fiduciary markers or inter-spacecraft communications capabilities that help the servicer in estimating their relative pose. In the latter case, the servicer has to estimate the pose from the sole information provided by its sensors. While cooperative RPOs have been the norm over the past, they do not suit with the requirements of future on-orbit servicing missions where the targets were not designed for cooperative operations. Since most of those RPOs involve known spacecraft, we study the case of an operation targeting an uncooperative spacecraft whose CAD model is available.
    
    Different sensors can be used to observe the target spacecraft~\cite{opromolla2017review}, such as LIDAR systems~\cite{opromolla2017pose,christian2013survey}, infrared~\cite{shi2015uncooperative,rondao2022chinet}, time-of-flight~\cite{martinez2017pose} or stereo cameras~\cite{pesce2017stereovision}. However, their cost, mass, bulkiness, and power consumption make them unsuitable for low-cost missions in Low Earth Orbit. Over the past few years, the community started considering single cameras as the cheapest solution to the spacecraft pose estimation problem. Indeed, due to their low mass, low cost, and low power consumption, they can easily be integrated in less expensive satellites. Estimating the 6D pose of a target spacecraft from a single image is a complex task. The orbital lighting conditions differ significantly from those encountered on Earth. Since there is no atmospheric diffusion, the captured images suffer from large contrasts between exposed and shadowed spacecraft parts. This is further aggravated by the specular materials of which the spacecraft is made up. As there is no dust contamination, the spacecraft's surfaces remain specular forever. In addition, a spacecraft is often nearly symmetrical so that the proposed solution should be able to resolve ambiguities from subtle features. Finally, the pose estimation pipeline must run on limited resources provided by space-grade hardware~\cite{cosmas2020fpga,leon2022fpga_asip}.
    
    Deep-learning based solutions~\cite{chen2019dlr,sharma2019spnv1,park2019krn} have been proposed to solve the spacecraft pose estimation problem but they are exclusively trained on synthetic images that hardly capture the illumination conditions encountered in Low-Earth Orbit and rely on a simplified spacecraft model. Due to these mismatches between the synthetic and real domains, those solutions encounter significant generalization issues. This domain gap problem, \ie, the mismatch between the synthetic training domain and the real test domain, has recently attracted a strong interest with the Spacecraft Pose Estimation Challenge (SPEC) 2021~\cite{park2023spec2021} hosted by the European Space Agency and Stanford University. Several works~\cite{perez2022spacecraft,wang2023bridging,park2023spec2021} address the domain gap by exploiting techniques such as Generative Adversarial Networks (GANs)~\cite{goodfellow2014generative,wang2023bridging}, adversarial training~\cite{park2023spec2021} or pseudo-labeling~\cite{perez2022spacecraft}. Although those works successfully bridge the gap towards a specific target domain, they rely on domain adaptation strategies, \ie, techniques that exploit priors on the target domain to bridge the domain gap. As a result, those techniques are unsuitable for real use cases where little information is known about the target domain. Unlike them, our work aims at bridging the domain gap through domain generalization, \ie, techniques that require no prior knowledge of the target and aim at enlarging the range of domains covered by the trained model.
    
    Our work addresses the domain generalization problem by introducing a learning strategy that relies on aggressive data augmentation policies, \ie, a form of Domain Randomization, and multi-task learning to prevent the network from overfitting on the training domain and enforce it to learn domain-invariant features. This strategy is applied on a novel, fully neural based, pose estimation network which embeds most of the salient components integrated in recent solutions proposed in the related literature. It relies on an intermediate detection of keypoints to recover the spacecraft relative 6D pose, decoupling the image processing step from the pose estimation process. We show that our method achieves State-of-The-Art accuracy without requiring on-orbit images at training.

\section{Related Works}
    The first methods proposed for spacecraft pose estimation, such as the Sharma-Ventura-D’Amico method~\cite{sharma2018robust}, only achieved accurate estimates with a high failure rate. Sharma and D'Amico~\cite{sharma2019spnv1} further introduced SPN, the first deep learning-based solution. The problem gained in interest with SPEC2019~\cite{kisantal2020spec2019}. Several methods emerged from SPEC2019. Chen \etal~\cite{chen2019dlr} won the challenge with an approach based on the prediction of pre-defined keypoints through a HRNet~\cite{sun2019hrnet} backbone. The pose was then computed by solving a non-linear least squares problem using the Levenberg–Marquardt algorithm~\cite{marquardt1963algorithm}. Park \etal~\cite{park2019krn} used a MobileNet-v2~\cite{sandler2018mobilenetv2} to regress keypoints, which were then used by a Perspective-n-Point (PnP) solver to predict the spacecraft pose. Similarly, Gerard \etal~\cite{gerard2019segmentation} predicted pre-defined keypoints through a DarkNet-53~\cite{redmon2018yolov3} and estimated the pose using EPnP~\cite{lepetit2009epnp} combined with RANSAC~\cite{fischler1981ransac}. WDR~\cite{hu2021wdr} improved this solution by considering a hierarchical architecture that predicts keypoints at different scales. Unlike all those works relying on a geometrical optimization problem, Proença and Gao~\cite{proencca2020ursonet} proposed an end-to-end solution, URSONet. They exploited a ResNet architecture~\cite{he2016residualblocks} to extract features which were then used to regress the spacecraft position and predict its relative orientation through probabilistic quaternion fitting. Following SPEC2019, several works addressed the problem through end-to-end approaches~\cite{sonawani2020assistive,legrand2022end} or keypoint-based ones~\cite{black2021realtime,cassinis2022validationdomainshift}. Some of them~\cite{posso2022mobile,carcagni2022lightweight} further focused on lightweight approaches to meet the complexity requirements of space-grade hardware. In this paper, we introduce a novel method that is both keypoint-based and end-to-end. We further consider lightweight networks to decrease its computational complexity.

    SPEC2019 also highlighted that all deep learning-based solutions overfit the synthetic training domain and therefore achieve a limited accuracy on real images. This problem, known as the domain gap, motivated a second edition, SPEC2021~\cite{park2023spec2021}, that brought several approaches to deal with the domain gap. They correspond either to domain adaptation or to domain generalization strategies. While the former assume that the target domain is known, the latter do not rely on any prior on this domain. As a result, \textit{domain adaptation techniques use images from the target domain} during training while \textit{domain generalization techniques only exploit synthetic images.} All top-scoring methods were unsurprisingly relying on domain adaptation techniques. The \textit{TangoUnchained} team~\cite{park2023spec2021} used data augmentation policies and adversarial landmark regression. Wang \etal~\cite{wang2023bridging} further explored the use of adversarial techniques by using Cycle-GAN~\cite{zhu2017unpaired} to train the network on semi-real images, and self-training. P{\'e}rez-Villar \etal~\cite{perez2022spacecraft} relied on pseudo-labeling of the HIL domains. Finally, Legrand \etal~\cite{legrand2024leveraging} exploited Neural Radiance Fields~\cite{mildenhall2021nerf} to augment the size of a training set made of few real images to train a pose estimator. Regarding the domain generalization strategies, Park and D'Amico~\cite{park2023robust} introduced SPNv2 which relies on multi-task learning and image augmentation policies to bridge the domain gap. In addition, they proposed an Online Domain Refinement (ODR) technique to fine-tune a pre-trained model on-board during the mission to achieve on-orbit domain adaptation. Park and D'Amico further improved their work with SPNv3~\cite{park2024spnv3} which considered additional data augmentations such as DeepAugment~\cite{hendrycks2021many} and RandConv~\cite{xu2020robust}, as well as transformer-based architectures~\cite{xu2022vitpose}. Ulmer \etal~\cite{ulmer20236d} relied on dense 2D-3D correspondences prediction combined with pose hypothesis and refinement as well as data augmentation techniques. Finally, Cassinis \etal~\cite{cassinis2023leveraging} exploited augmentation techniques to train a HRNet~\cite{sun2019hrnet} to perform keypoint regression. This paper bridges the domain gap through a learning strategy that aims at domain generalization, \ie, assuming no prior knowledge on the target domain. Our method builds on the domain generalization paradigm followed by SPNv2~\cite{park2023robust} and Cassinis \etal~\cite{cassinis2023leveraging} but offers a different perspective from those works. While the main purpose of the multi-task learning framework used to train the backbone of SPNv2 is to promote task-agnostic features, our framework further encourages the learning of a geometry-aware representation through the auxiliary face segmentation task. Similarly, while the Data Augmentation policies of SPNv2 and Cassinis \etal aimed at enlarging the training set distribution by synthesizing plausible samples, our Domain Randomization strategy goes one step further. Our augmentation policies are designed to aggressively transform the original images through, e.g., deletion of  parts of the images, simulation of severe over/under-exposure, or texture alterations. This forces the network to exploit more subtle features. 

\section{Method}
\label{sec_method}
        This section describes our method for enlarging the range of image domains properly managed when estimating the target 6D pose, \ie, position and orientation. It relies on a novel architecture as well as a domain generalization learning strategy. \Cref{sec_arch} describes the network architecture while the domain generalization strategy is developed in \Cref{sec_multi_task}, which introduces our Multi-Task Learning (MTL) framework, and \Cref{sec_domain_randomization}, which presents our domain randomization strategy.

\begin{figure*}[b]
    \addtocounter{figure}{1} 
    \centering
    \includegraphics[width=0.75\linewidth]{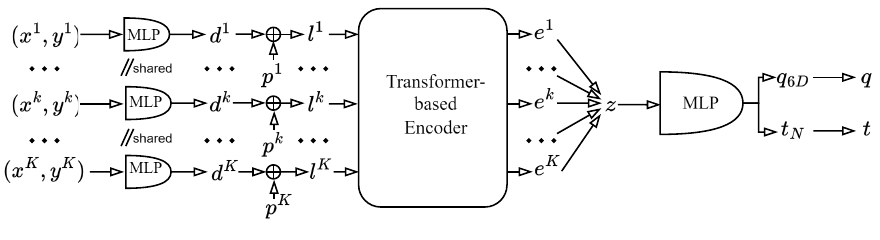}
    \caption{\label{fig_posemodule} Overview of our Pose Estimation Model which relies on a transformer-based encoder to predict the target 6D pose from the coordinates of K pre-defined keypoints.}
\end{figure*} 

\subsection{Network Architecture}
    \label{sec_arch}
    The input image depicts a spacecraft at a variable distance from the camera. As a result, the spacecraft may occupy a substantial portion of the image or only a small part. To achieve scale invariance and limit the computational complexity of the method, only a crop of the original image centered on the spacecraft is provided to the subsequent network. We assume that a bounding box centered on the target is available, \eg, through an object detection network~\cite{chen2019dlr,park2019krn} or using a pose estimate obtained at a previous time step. This bounding box is used to crop the image and resize it to a resolution $W \times H$. As depicted in \Cref{fig:network_overview}, this image is processed through a Keypoint Positioning Network (KPN) to predict the coordinates of $K$ pre-defined keypoints. Those coordinates are then processed by a Pose Estimation Model (PEM) which regresses the 6D pose through an attention-based encoder. This method decouples the image processing, \ie, keypoint regression, from the pose estimation, thereby primarily restricting the generalization issue to the keypoint estimation module and leveraging the pose estimation robustness to deal with erroneous keypoints. \Cref{sec_keypoints} describes the KPN architecture while the Pose Estimation Model is presented in \Cref{sec_posemodule}.

\begin{figure}[t]
    \setcounter{figure}{0} 
    \centering
    \includegraphics[width=0.9\linewidth]{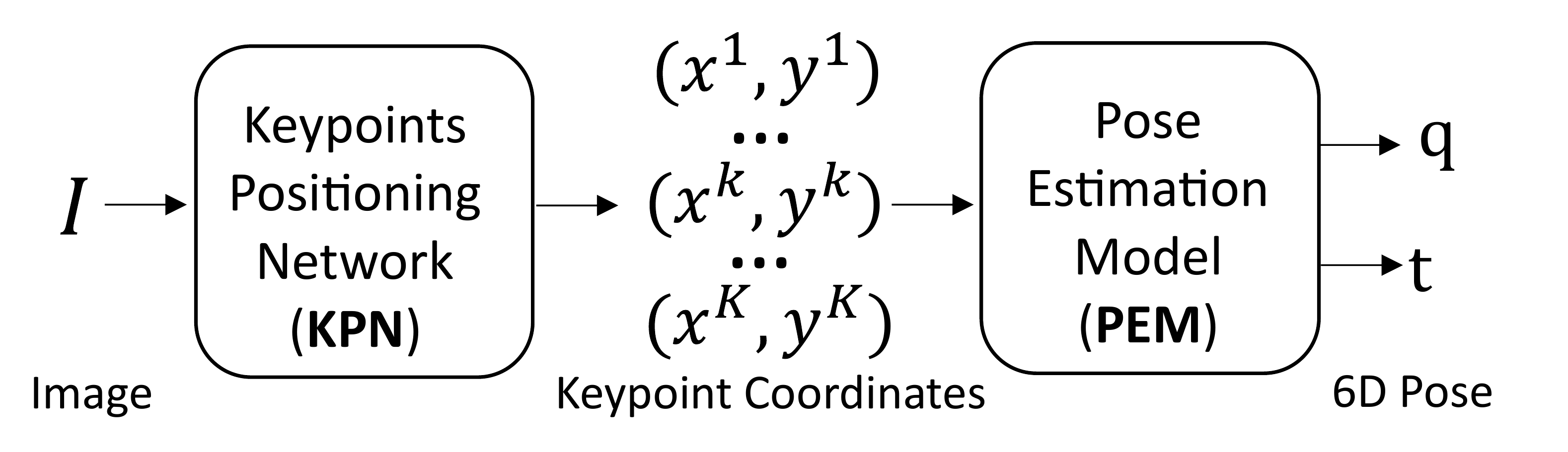}
    \caption{\label{fig:network_overview}Architecture overview. The target 6D pose is predicted by the Pose Estimation Model (PEM) trough pre-defined keypoints coordinates regressed from the input image by the Keypoint Positioning Network (KPN).}
    \setcounter{figure}{2} 
\end{figure}

    \subsubsection{Keypoint Positioning Network} 
    \label{sec_keypoints}
        The Keypoint Positioning Network (KPN) aims at estimating the 2D coordinates of $K$ pre-defined keypoints from the image. For this purpose, we rely on a first step of histogram equalization to bring the histograms of all the images, \ie, training and testing, closer to a uniform distribution which results in a reduced gap between the domains. \Cref{sec_exp_ablation_Equalization} provides an ablation study that demonstrates the interest of this equalization on the network generalization abilities. 
        
        The KPN relies on a high-resolution network backbone, \ie, HRNet~\cite{sun2019hrnet} that predicts feature maps at multiple resolutions. Those multi-resolution maps are then aggregated into $K$ heatmaps of size ($W/4$, $H/4$) using a 4-layers Bi-directional Feature Pyramid Network (BiFPN)~\cite{tan2020efficientdet}. The normalized keypoint coordinates ($x^k_N$,$y^k_N$) are then obtained by applying a DSNT~\cite{nibali2018dsnt}, \ie, Differentiable Spatial to Numerical Transform, to each heatmap $h^k$. The DSNT is used because it achieves a better prediction accuracy and is differentiable. The normalized $x$ and $y$ coordinates of the $k^{th}$ keypoint are then computed as follows,
        \begin{equation}
            x^k_N = \sum_{i=1}^{W/4} \sum_{j=1}^{H/4} h_N^k(i,j) x(i,j),
        \end{equation}
        and, 
        \begin{equation}
           y^k_N = \sum_{i=1}^{W/4} \sum_{j=1}^{H/4} h_N^k(i,j) y(i,j),
        \end{equation}
        where $x(i,j)$ and $y(i,j)$ are the $x$ and $y$ normalized coordinates at location $(i,j)$, while $h_N^k$ denotes the heatmap $h^k$ normalized so that each heatmap sums up to $1$.\\
        Finally, the keypoint coordinates in the full-resolution input image ($x^k$,$y^k$) are computed as
        \begin{equation}
            ( x^k , y^k ) = ( x_0 + x^k_N w , y_0 + y^k_N h ) ,
        \end{equation}
        where $[x_0, y_0, w, h]$ comes from the bounding box used to crop the spacecraft from the full resolution image.

        The KPN is trained by minimizing the average Euclidean distance between the predicted and ground-truth normalized keypoint coordinates, $(x_{N}^{k},y_{N}^{k})$ and $(\hat{x}_{N}^{k},\hat{y}_{N}^{k})$, respectively, \ie, 
        \begin{equation}
            L_{Kpts} = \frac{1}{K} \sum_{k=1}^{K} \sqrt{(x_{N}^{k}-\hat{x}_{N}^{k})^{2} + (y_{N}^{k}-\hat{y}_{N}^{k}})^{2}.
            \label{eq_kpts_loss}
        \end{equation}

        \subsubsection{Pose Estimation Model}
        \label{sec_posemodule}
    
        As illustrated in \Cref{fig_posemodule}, the Pose Estimation Model (PEM) aims at predicting the spacecraft 6D pose, \ie, position $t$ and orientation $q$, from the 2D coordinates of the $K$ pre-defined keypoints estimated by the KPN. First, to express the input coordinates in a representation that is more suited to the pose estimation task, each of those $K$ 2D coordinates, ($x^k$,$y^k$), is normalized and mapped to a higher dimension through a Multi-Layer Perceptron (MLP) resulting in an embedding $d^k$ of dimension D. Then, for each keypoint, a learned positional embedding $p^k$ is added to $d^k$ so that the resulting embedding, $l^k$, implicitly contains information on which physical keypoint is represented. The resulting $K$ embeddings are fed into the encoder of a transformer which outputs $K$ embeddings of the same dimension, which are concatenated into a single vector that is fed in a MLP made of three hidden layers with 256 neurons per layer. It outputs the normalized 3D position of the target in the camera frame as well as a 6D vector representing the rotation that aligns the spacecraft inertial frame to the camera frame. Predicting the rotation as a 6D vector has been inspired by Zhou \etal~\cite{zhou2019continuity}, who demonstrated that regressing rotations represented by Euler angles, axis-angle or quaternions was inefficient due to their discontinuous representation of the SO(3) space. At inference, the translation $t$ in meters is computed from the normalized regressed translation vector while the 6D-vector representing the rotation is mapped to the corresponding quaternion $q$. A salient specificity of our architecture lies in the transformer-based encoder. Its purpose is to identify, through self-attention layers, consistent patterns in the input embeddings, thereby providing outliers-resilience capabilities to the model. \Cref{sec_exp_ablation_Architecture} provides an ablation study on the PEM architecture.   
        
        The Pose Estimation Model is trained on top of the KPN. In this phase, only the PEM weights are optimized by minimizing the PEM loss, \ie, $L_{PEM}$, which combines the normalized translation error with the error between the ground-truth and predicted 6D representations of the rotations, \ie,
        \begin{equation}
            L_{PEM} = \frac{\left | \left | t - \hat{t} \right | \right |_2}{\left | \left | t \right | \right |_2} + \left | \left | q_{6D} - \hat{q}_{6D} \right | \right |_1,
        \end{equation}
        where $t$ and $\hat{t}$ are the predicted and ground-truth positions, $q_{6D}$ and $\hat{q}_{6D}$ are the 6D representations of the predicted and ground-truth rotations.

    \subsection{Multi-Task Learning}
    \label{sec_multi_task}
\begin{figure}[b]
    \centering
    \includegraphics[width=0.99\linewidth]{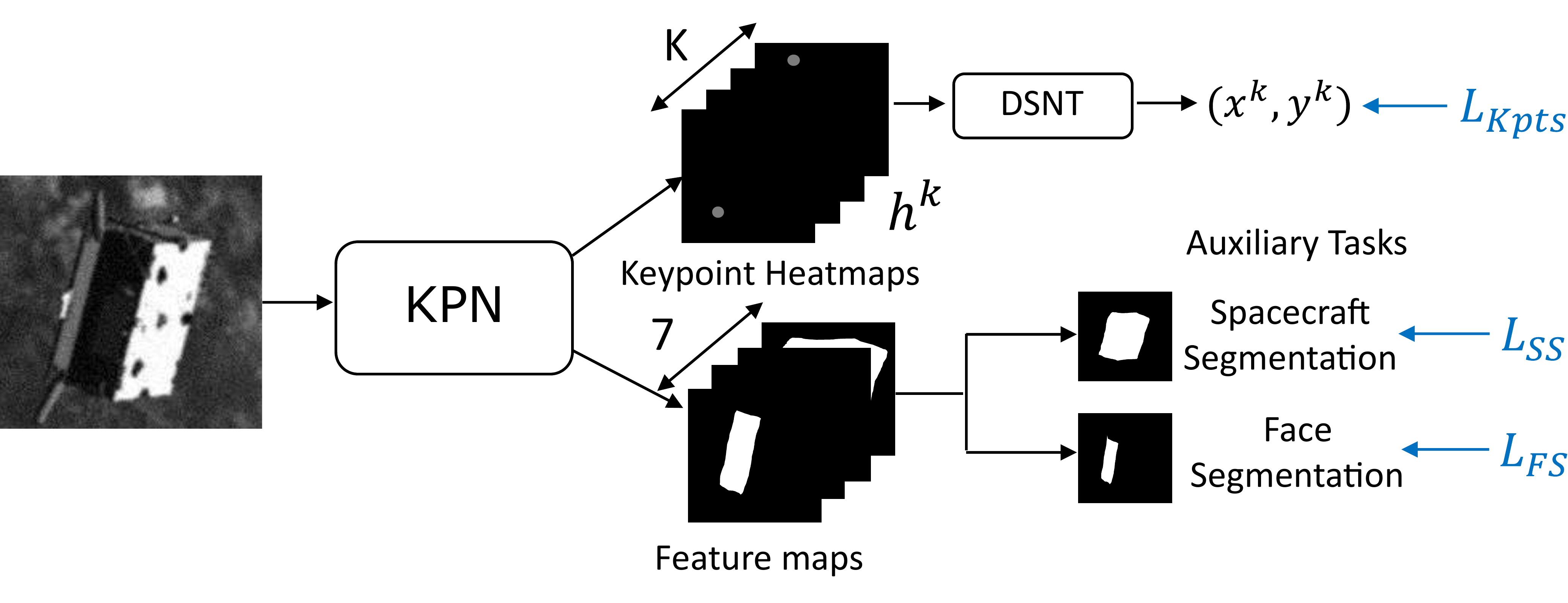}
    \caption{\label{fig_mtl} Using our MTL framework, the network is supervised on the regression of keypoint coordinates and on two auxiliary tasks that aim at segmenting the spacecraft and each of its faces.}
\end{figure}

        Inspired by previous works~\cite{park2023robust,park2023spec2021} in domain generalization, our training procedure relies on multi-task learning to improve the KPN generalization abilities. In a multi-task framework, the learning of a main task is leveraged by learning auxiliary tasks in parallel with the main one. In our method, in addition to the main task of regressing keypoint coordinates, two segmentation tasks are considered. The first one aims at segmenting the spacecraft while the second one aims at segmenting each of the six faces of the spacecraft body. This multi-task learning strategy is illustrated in \Cref{fig_mtl}, where the KPN outputs $K$ keypoint heatmaps and seven feature maps used by the auxiliary segmentation tasks.            

        In addition to the loss on the keypoints (see \Cref{eq_kpts_loss}), the KPN is therefore trained on both auxiliary tasks through Binary Cross Entropy losses~\cite{good1952rational}, \ie, $L_{SS}$ and $L_{FS}$. The total loss back-propagated through the KPN, $L_{KPN}$, therefore sums up the three losses weighted by two hyper-parameters, $\beta_{Kpts}$ and $\beta_{Multi}$, that determine the impact that the auxiliary tasks should have on the KPN training, \ie,
        \begin{equation}
            L_{KPN} = \beta_{Kpts} L_{Kpts} + \beta_{Multi} ( L_{SS} + L_{FS} ).\label{equation_kpn}
        \end{equation}

\begin{figure}[t]
    \centering
    \setlength\tabcolsep{0.0pt} 
    \begin{tabular}{cccc}
        \includegraphics[width=0.25\linewidth]{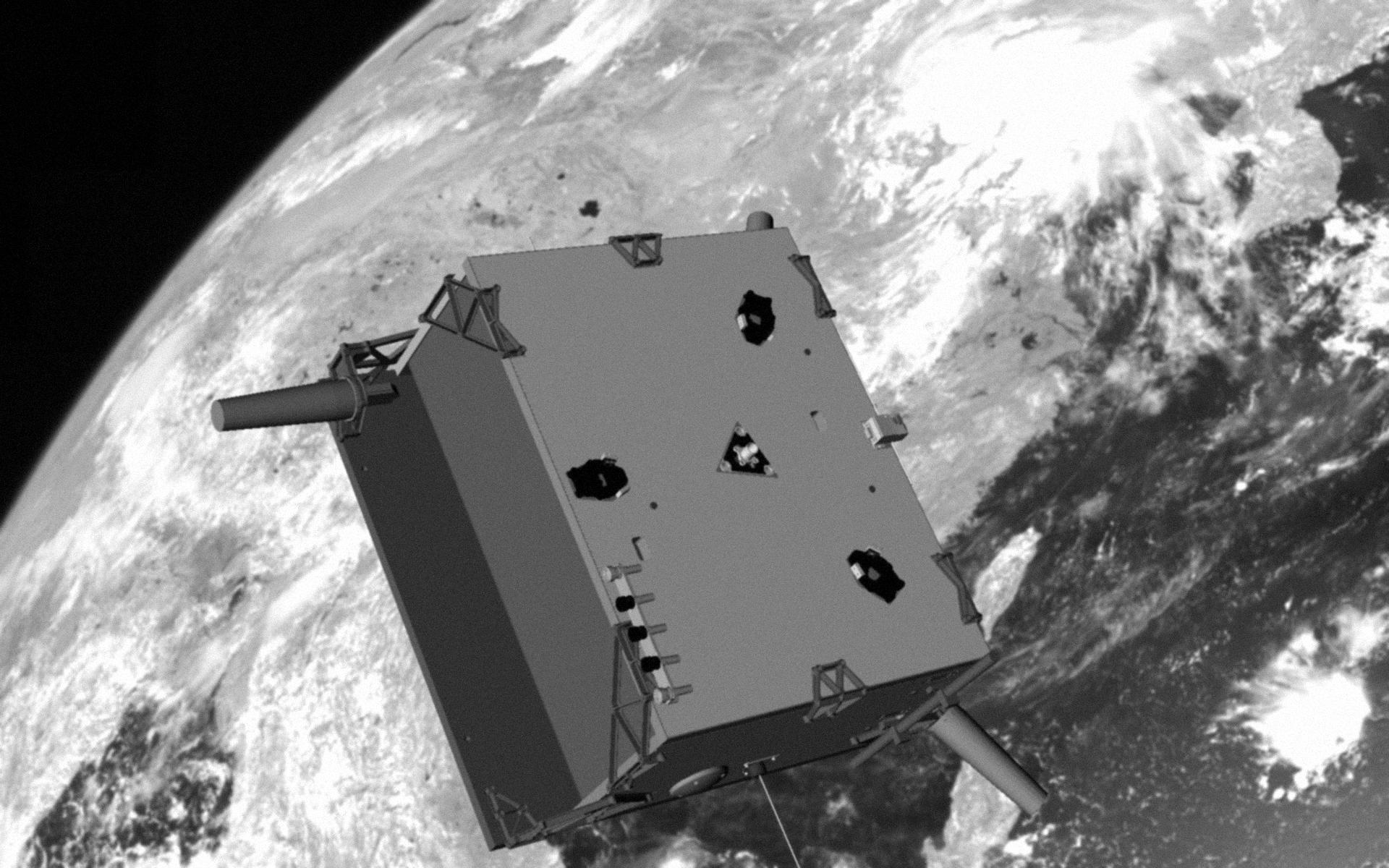} &
        \includegraphics[width=0.25\linewidth]{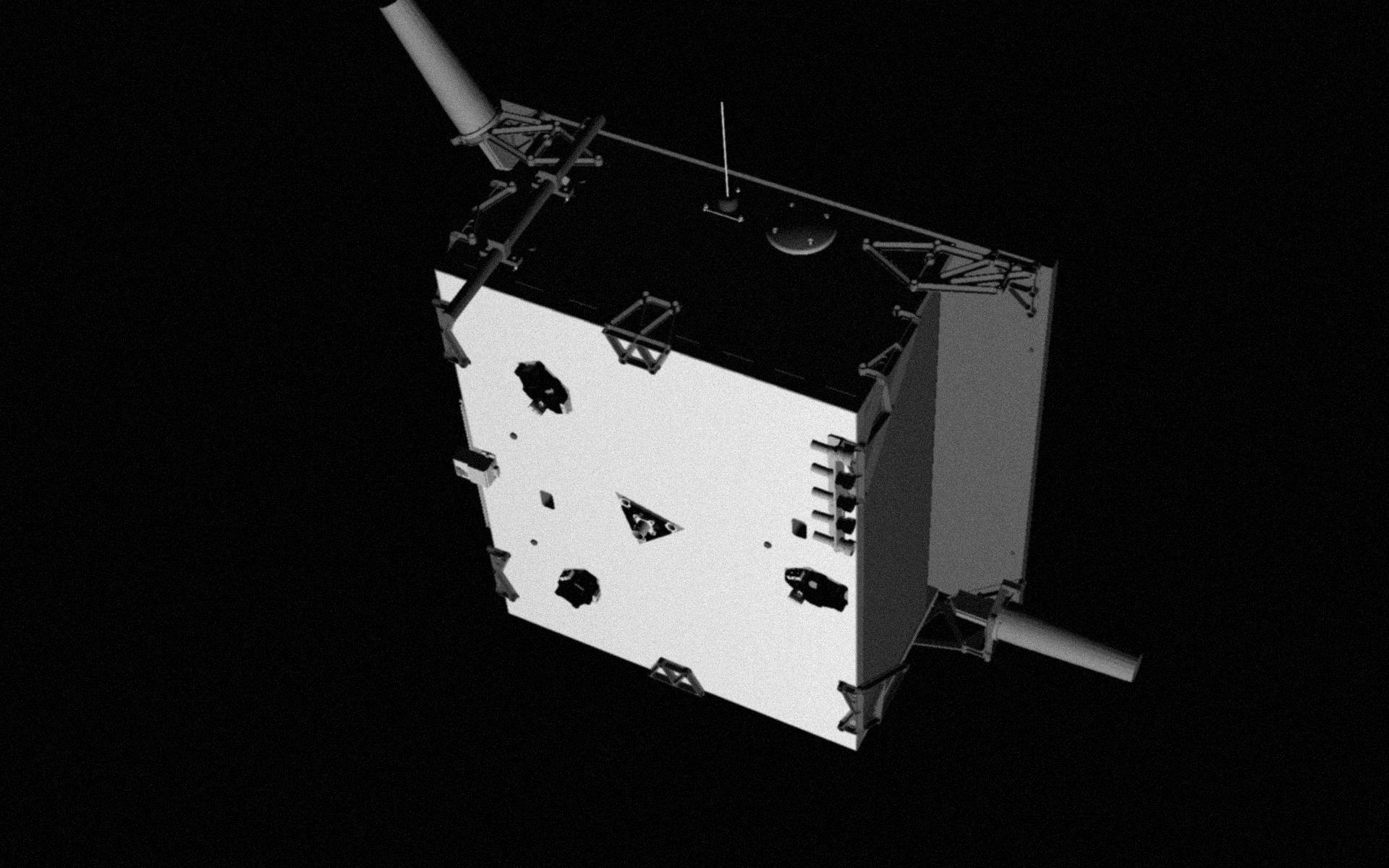} & 
        \includegraphics[width=0.25\linewidth]{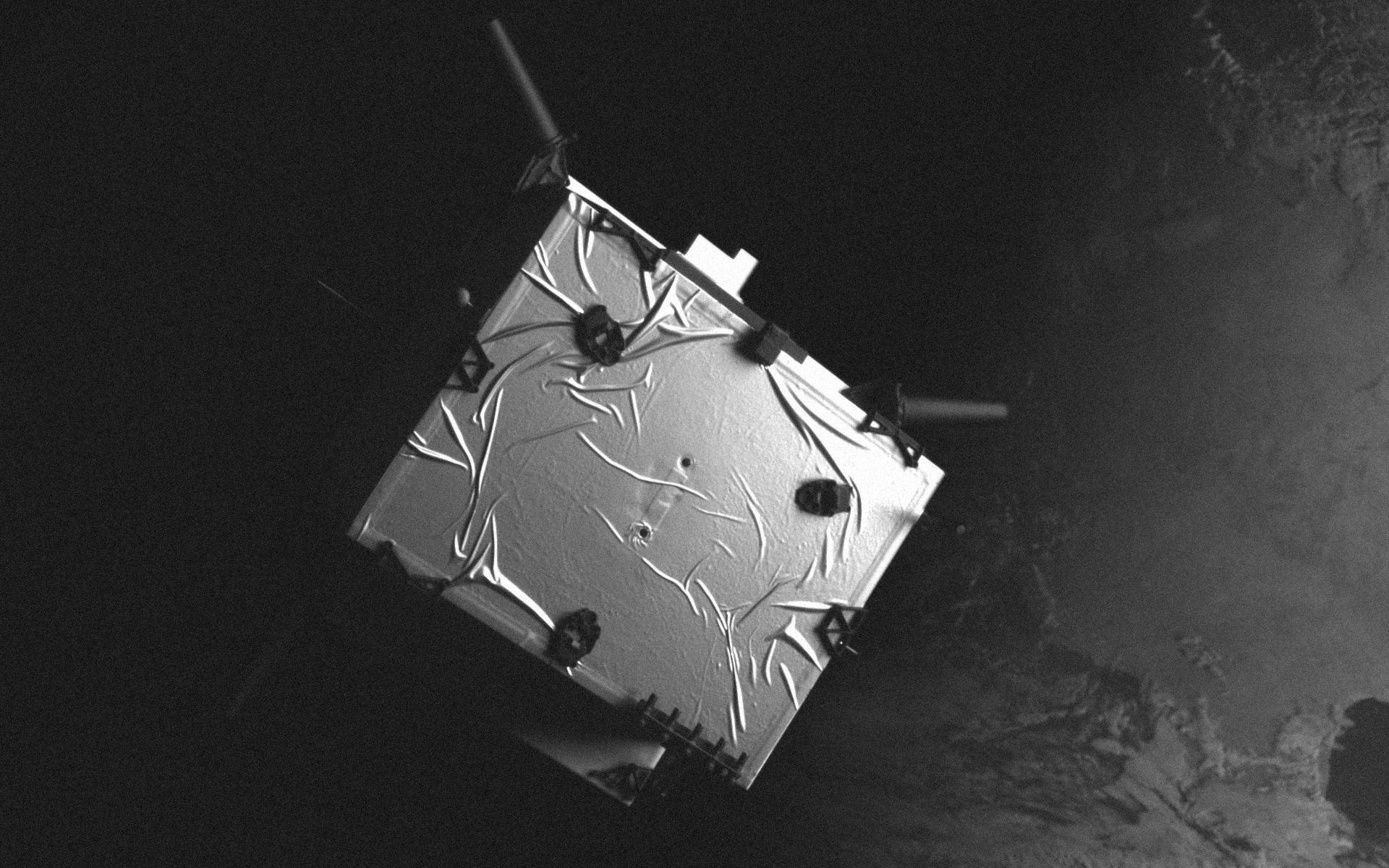} & 
        \includegraphics[width=0.25\linewidth]{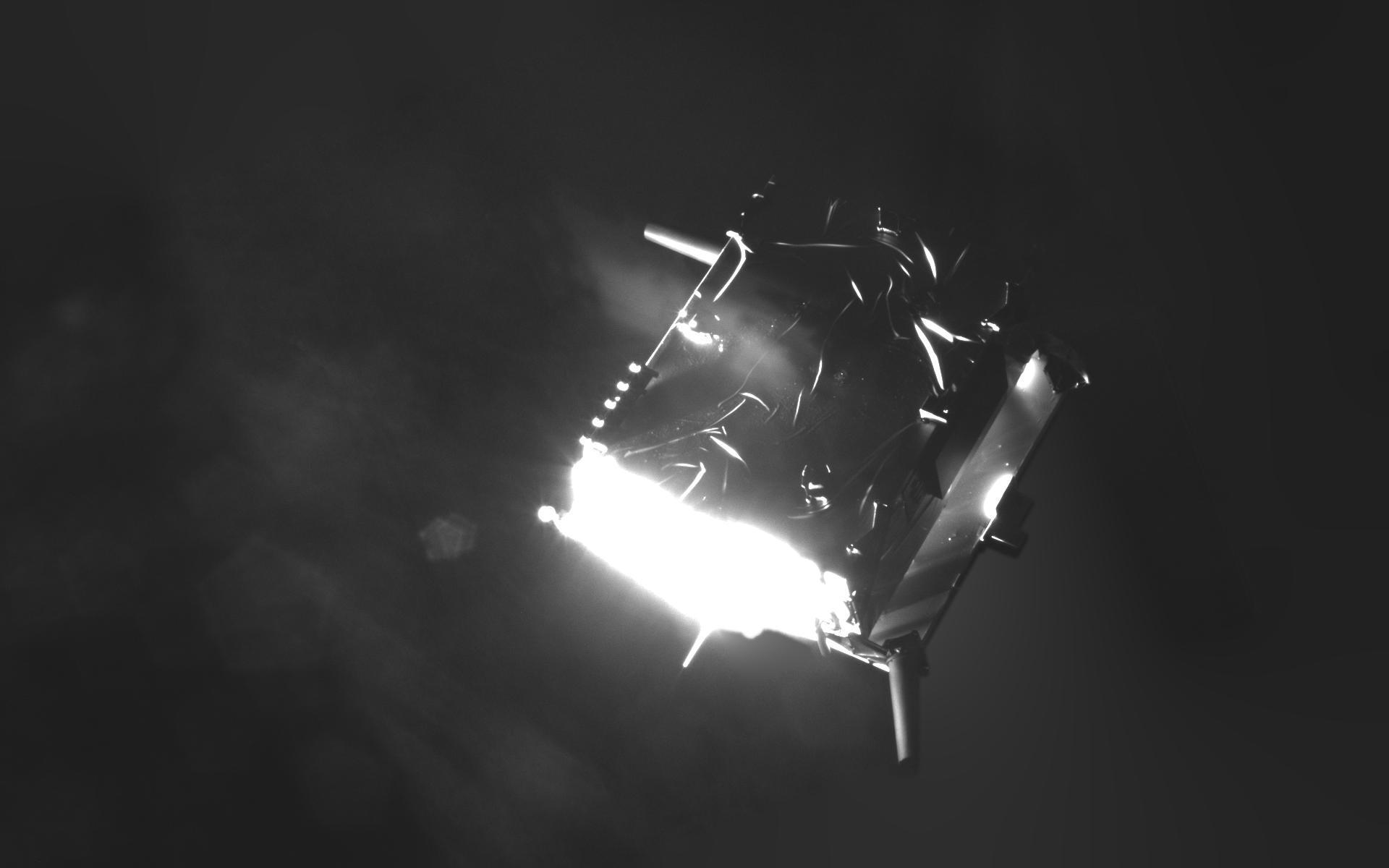}
    \end{tabular}
    \caption{\label{fig_MLI} SPEED+~\cite{park2022speed+} synthetic (left) and HIL images (right). HIL images exhibit highly textured surfaces, due to the Multi-Layer Insulation (MLI), that are not properly rendered in the synthetic images.}
\end{figure}

    \subsection{Domain Randomization}
    \label{sec_domain_randomization}
    
        The origin of the domain gap comes from the two main limitations of the image rendering tool used to generate the training set. Firstly, the illumination conditions encountered on-orbit are not captured by the rendering tool. Indeed, the specular spacecraft parts reflect the oncoming light which results in over-exposed images or image artifacts such as lens flares. Furthermore, due to the lack of atmospheric diffusion, the spacecraft parts that are not directly illuminated cannot be seen. However, those over and under exposure impairments are poorly captured by the rendering tool.
        Secondly, the CAD model used for generating the training set suffers from mismatches with the actual spacecraft. For example, in the SPEED+ dataset~\cite{park2022speed+}, the Multi-Layer Insulation (MLI) represented in the CAD model is over-simplified compared to the actual one, as illustrated in \Cref{fig_MLI}. These illumination and CAD mismatches are responsible for the domain gap problem. Hence, we introduce a domain randomization~\cite{tobin2017domain} technique that aims at randomizing the training domain through an aggressive data augmentation strategy.

        Several techniques are considered to achieve invariance against the illumination conditions in addition to the classical Gaussian noise and Brightness \& Contrast augmentations. Hide\&Seek~\cite{singh2018hide} is used to deal with the under-exposed spacecraft parts by randomly erasing parts of the image while an exposure augmentation technique, inspired from PostLamp~\cite{sakkos2019illumination}, is applied on multiple randomly selected points in the image in order to improve the network ability to cope with over-exposed spacecraft parts.

\begin{figure*}[t]
    \centering
    \includegraphics[width=0.99\linewidth]{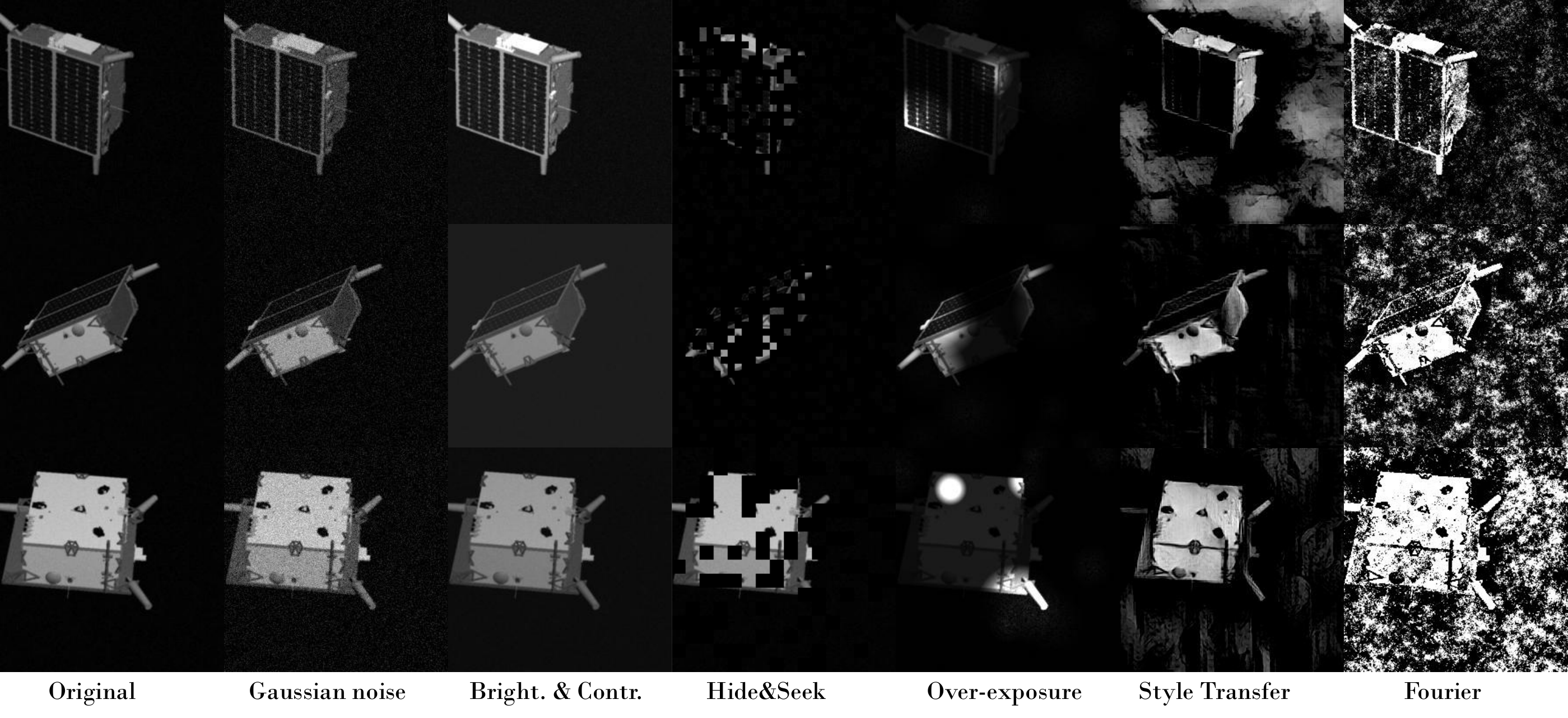}\\
    \caption{\label{fig_Single_DA_examples} SPEED+~\cite{park2022speed+} synthetic images and their augmented versions, considering the different data augmentation techniques used in this paper.} 
\end{figure*}

        In addition, since Convolutional Neural Networks tend to overfit on the image texture~\cite{geirhos2018imagenet} and since this texture is not properly rendered in synthetic images due to CAD mismatches, we introduce a texture augmentation technique that consists in applying either Neural Style Transfer~\cite{jackson2019style} or a Fourier-based augmentation. This Fourier-based augmentation is inspired by previous arts~\cite{xu2021fourier,xu2023fourier} and consists in computing the Fast Fourier Transform (FFT)~\cite{cooley1965algorithm} of the image and adding noise to its magnitude while preserving its phase. The augmented image is then reconstructed from the noisy magnitude and the preserved phase. Since the FFT phase is intact, the image semantic, \ie, the spacecraft pose and shape, is conserved while its aspect is corrupted, resulting in a texture-augmented image. \Cref{fig_Single_DA_examples} depicts some images from SPEED+ along their counterpart augmented with each of the considered data augmentations.

\begin{figure}[t]
    \centering
    \setlength\tabcolsep{0.0pt} 
    \renewcommand{\arraystretch}{0.0}
    \begin{tabular}{cccc}
        \includegraphics[width=0.24\linewidth]{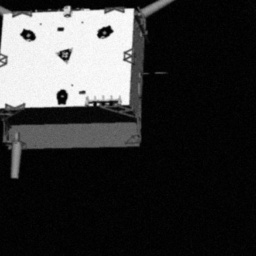} & 
        \includegraphics[width=0.24\linewidth]{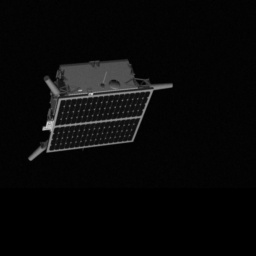} & \includegraphics[width=0.24\linewidth]{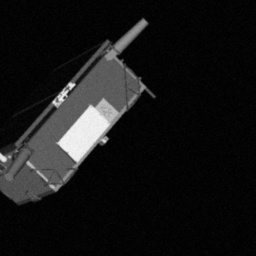} &
        \includegraphics[width=0.24\linewidth]{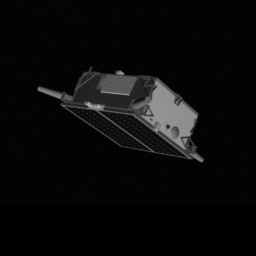} \\
        \includegraphics[width=0.24\linewidth]{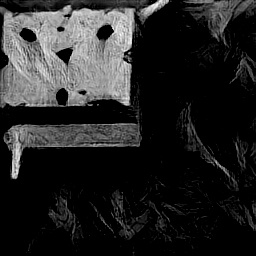} & 
        \includegraphics[width=0.24\linewidth]{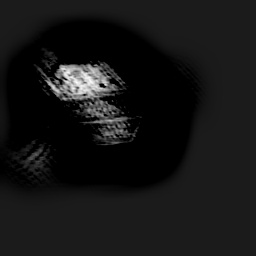} & \includegraphics[width=0.24\linewidth]{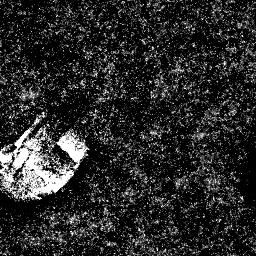} &
        \includegraphics[width=0.24\linewidth]{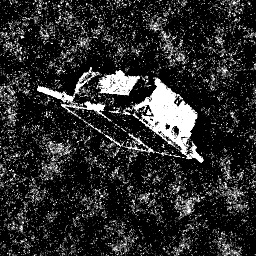}\\
    \end{tabular}
    \caption{\label{fig_Domain_Randomization_examples} Examples of our domain randomization technique. Original (top) and augmented (bottom) images from SPEED+~\cite{park2022speed+}. The image semantic is preserved but the texture and illumination significantly differ.}
\end{figure}

        Finally, the domain randomization is achieved by applying the augmentation techniques in an approach similar to RandAugment~\cite{cubuk2020randaugment}. We define a set of transforms composed of 6 data augmentation policies, \ie, (i) Gaussian noise, (ii) Brightness \& Contrast, (iii) Hide\&Seek, (iv) Exposure, (v) Texture and (vi) no augmentation. For each image of the train set, at each epoch, we randomly pick three different augmentation policies from the set and apply them consecutively. As illustrated in \Cref{fig_Domain_Randomization_examples}, this results in significantly randomized training images that preserve the semantic of the training set while significantly enlarging its distribution, consequently increasing the network generalization abilities.

 \begin{table*}[t]
    \caption{\label{tab_datasets} Overview of the SPEED+ dataset~\cite{park2022speed+}. \textit{Sunlamp} and \textit{Lightbox} are two Hardware-In-the-Loop (HIL) sets which are used only for validation. Because of the direct illumination of the Sun, which causes over-exposed images, \textit{Sunlamp} is the most challenging test set. The main differences between the HIL images and the synthetic ones are the illumination conditions and the spacecraft texture. \vspace{0.2cm}}
    \centering
    \setlength\tabcolsep{0pt} 
    \renewcommand{\arraystretch}{0.0}
    \begin{tabular}{|c|cc|cc|cc|}
        \hline
        \rule{0pt}{10pt}
        SPEED + & \multicolumn{2}{c|}{Synthetic} & \multicolumn{2}{c|}{\textit{Lightbox}}  & \multicolumn{2}{c|}{\textit{Sunlamp}}\\
        \hline
        \rule{0pt}{10pt}
        Domain (Use) & \multicolumn{2}{c|}{Synthetic (Train)} & \multicolumn{4}{c|}{HIL (Test)}\\
        \hline
        \rule{0pt}{10pt}
        Illumination & \multicolumn{2}{c|}{Synthetic} & \multicolumn{2}{c|}{Diffuse} & \multicolumn{2}{c|}{Direct} \\
        \hline
            Examples & \includegraphics[width=.14\linewidth]{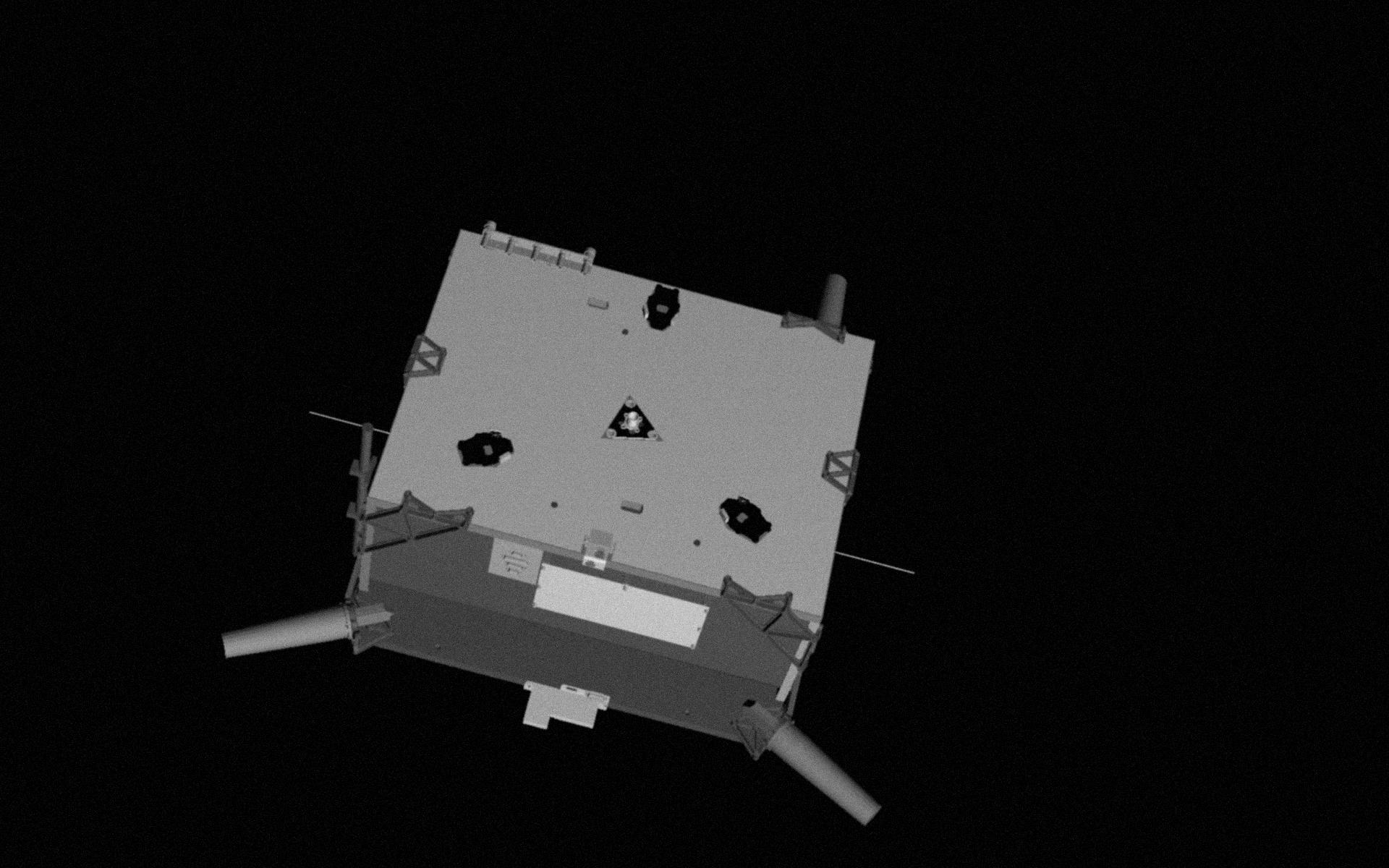} & \includegraphics[width=.14\linewidth]{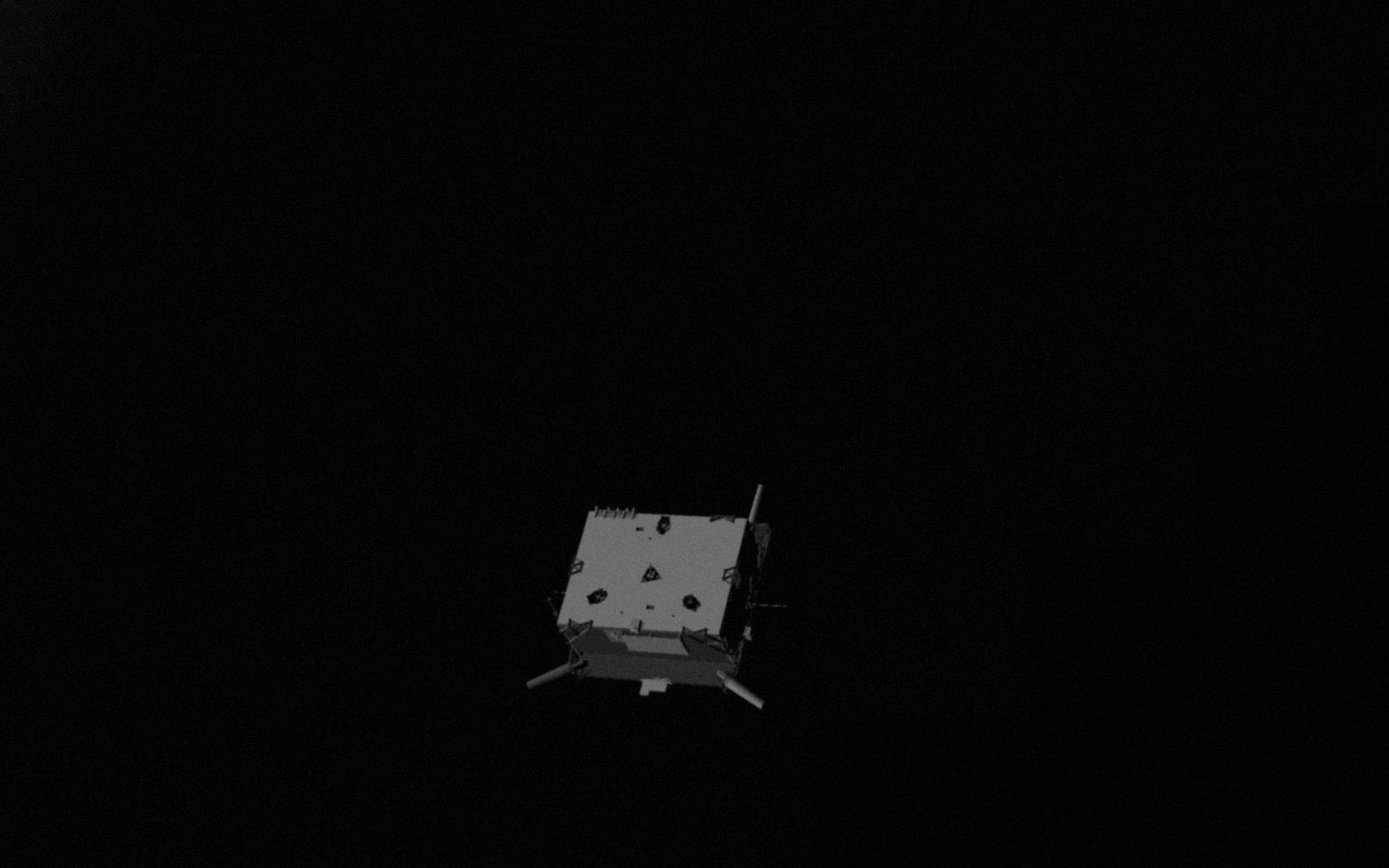} & \includegraphics[width=.14\linewidth]{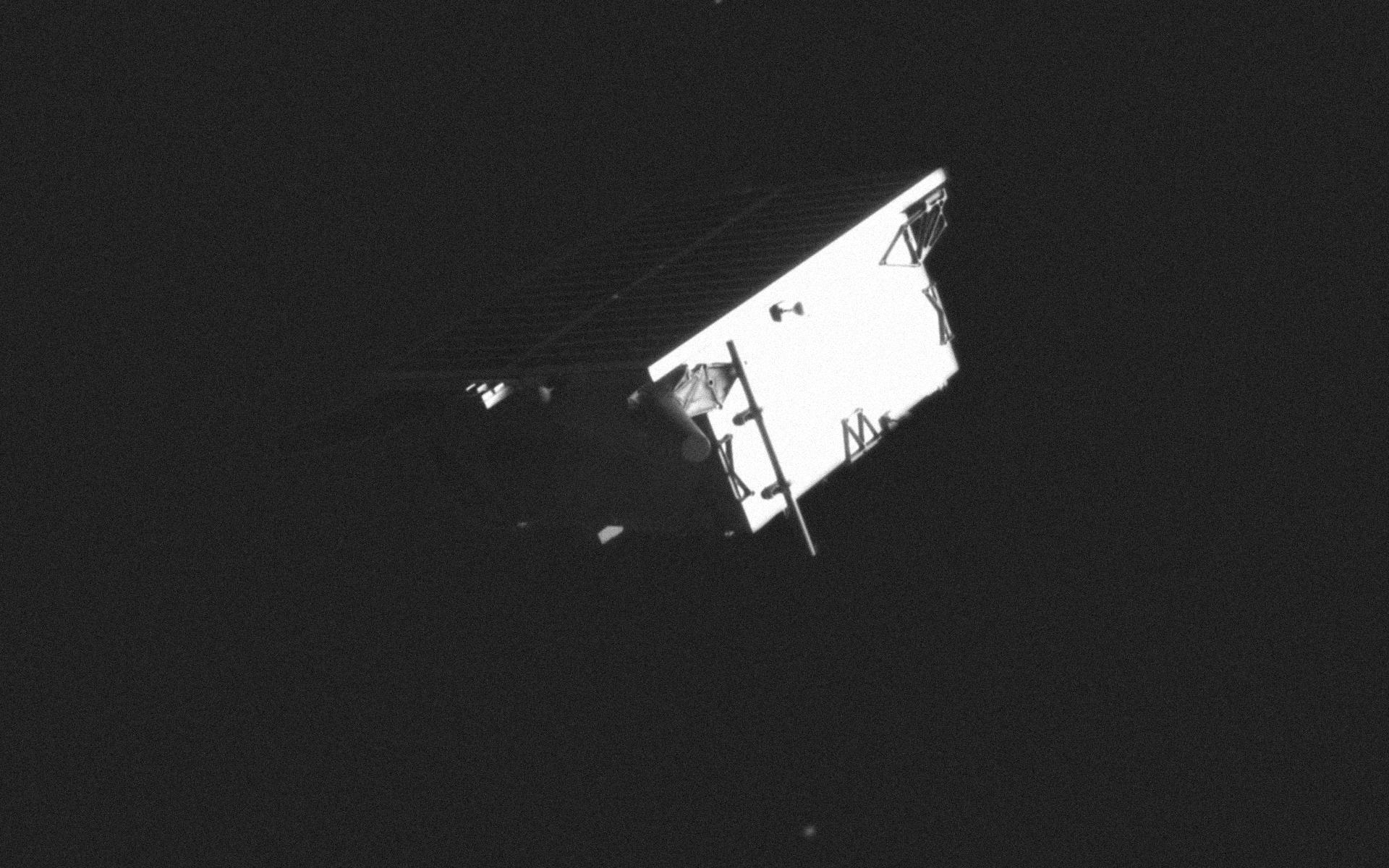} & \includegraphics[width=.14\linewidth]{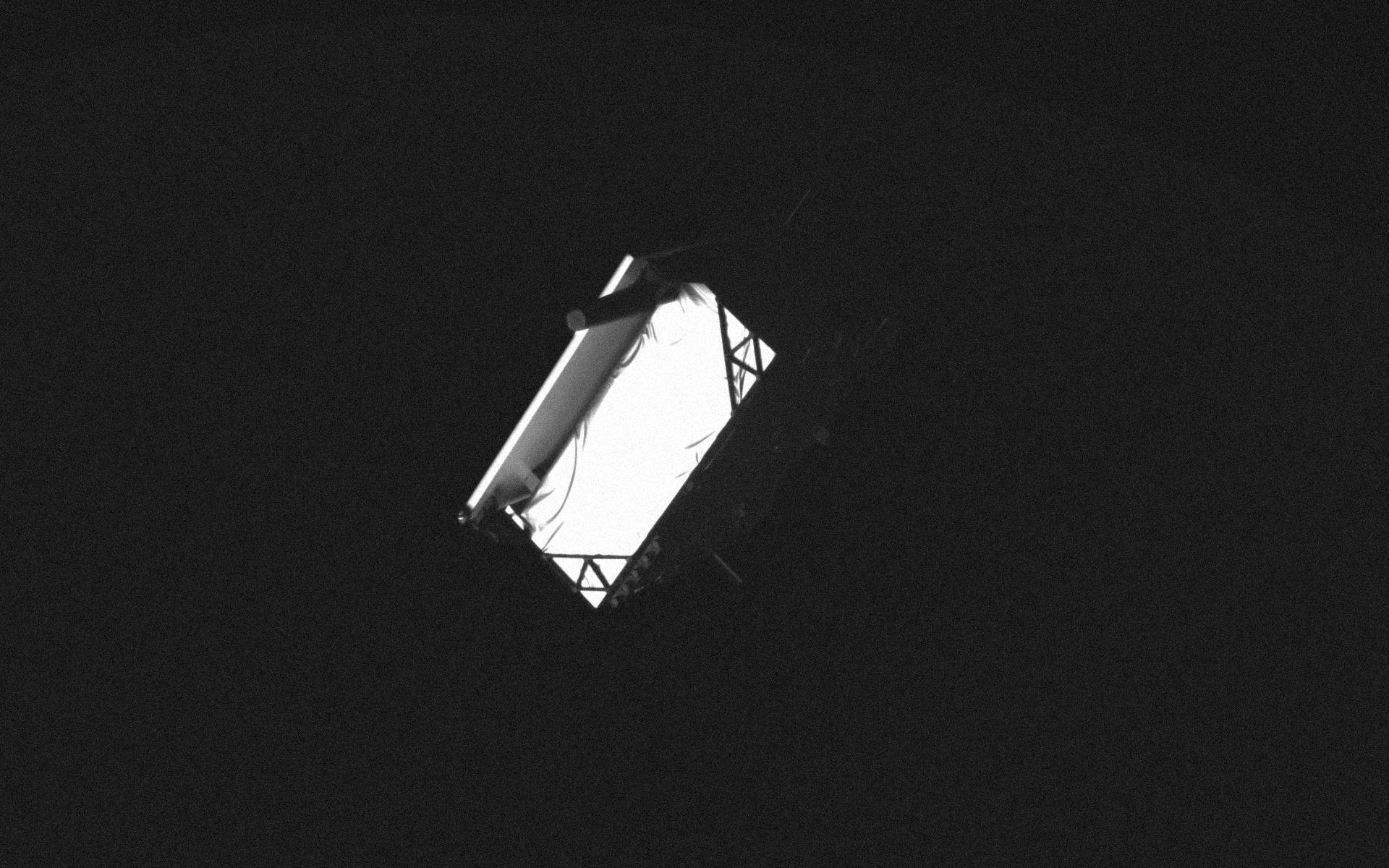} & \includegraphics[width=.14\linewidth]{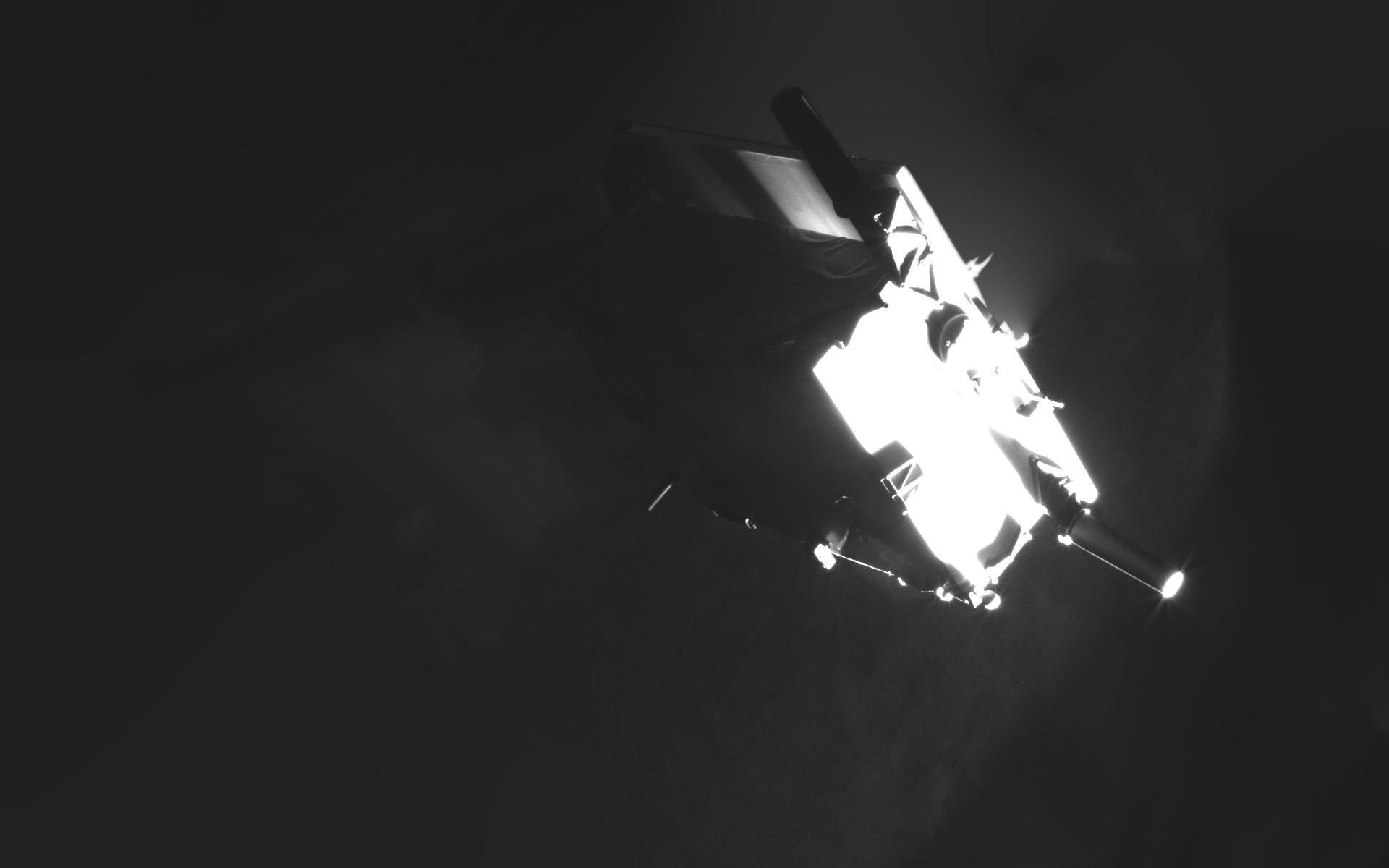} & \includegraphics[width=.14\linewidth]{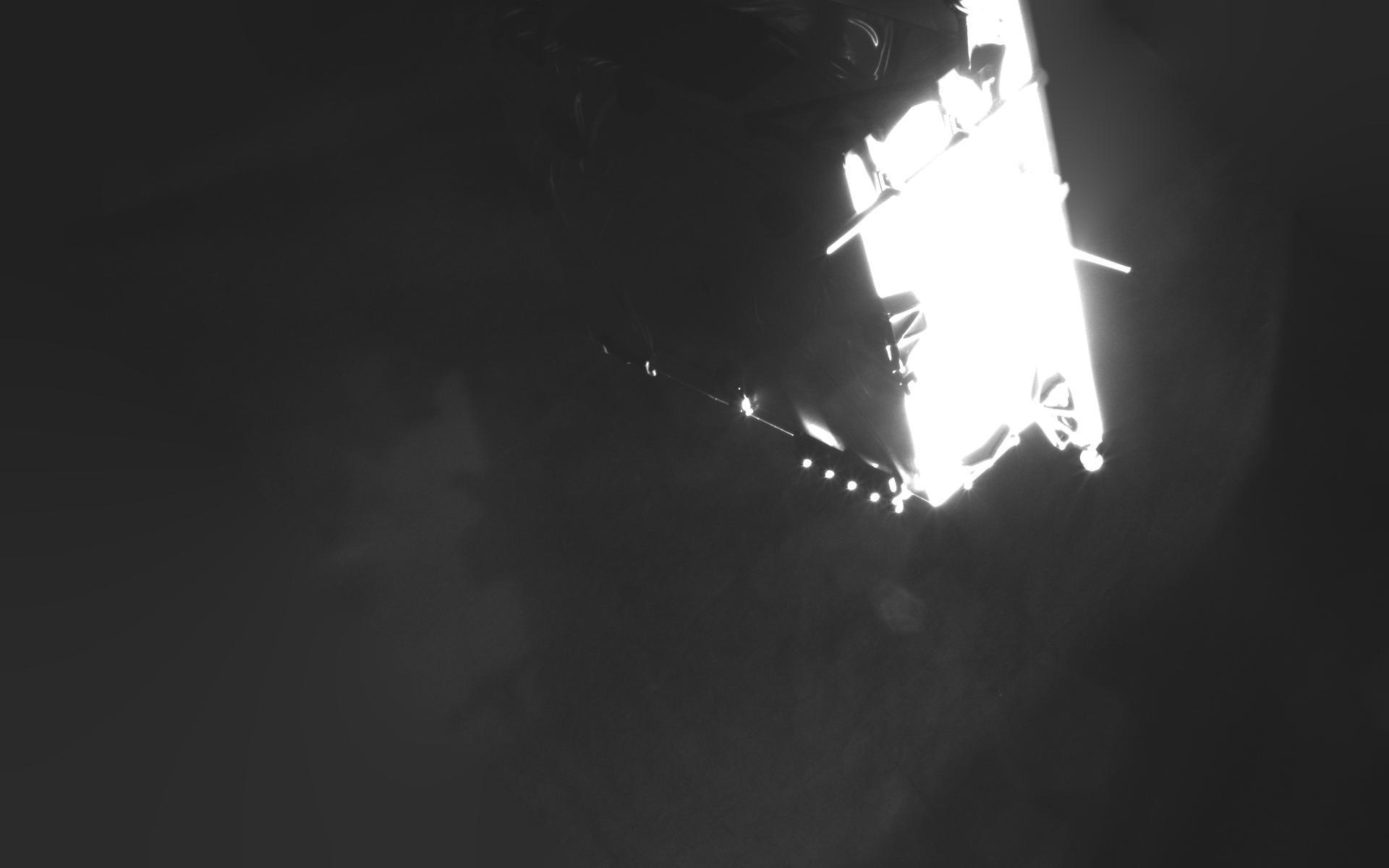}\\ 
            & \includegraphics[width=.14\linewidth]{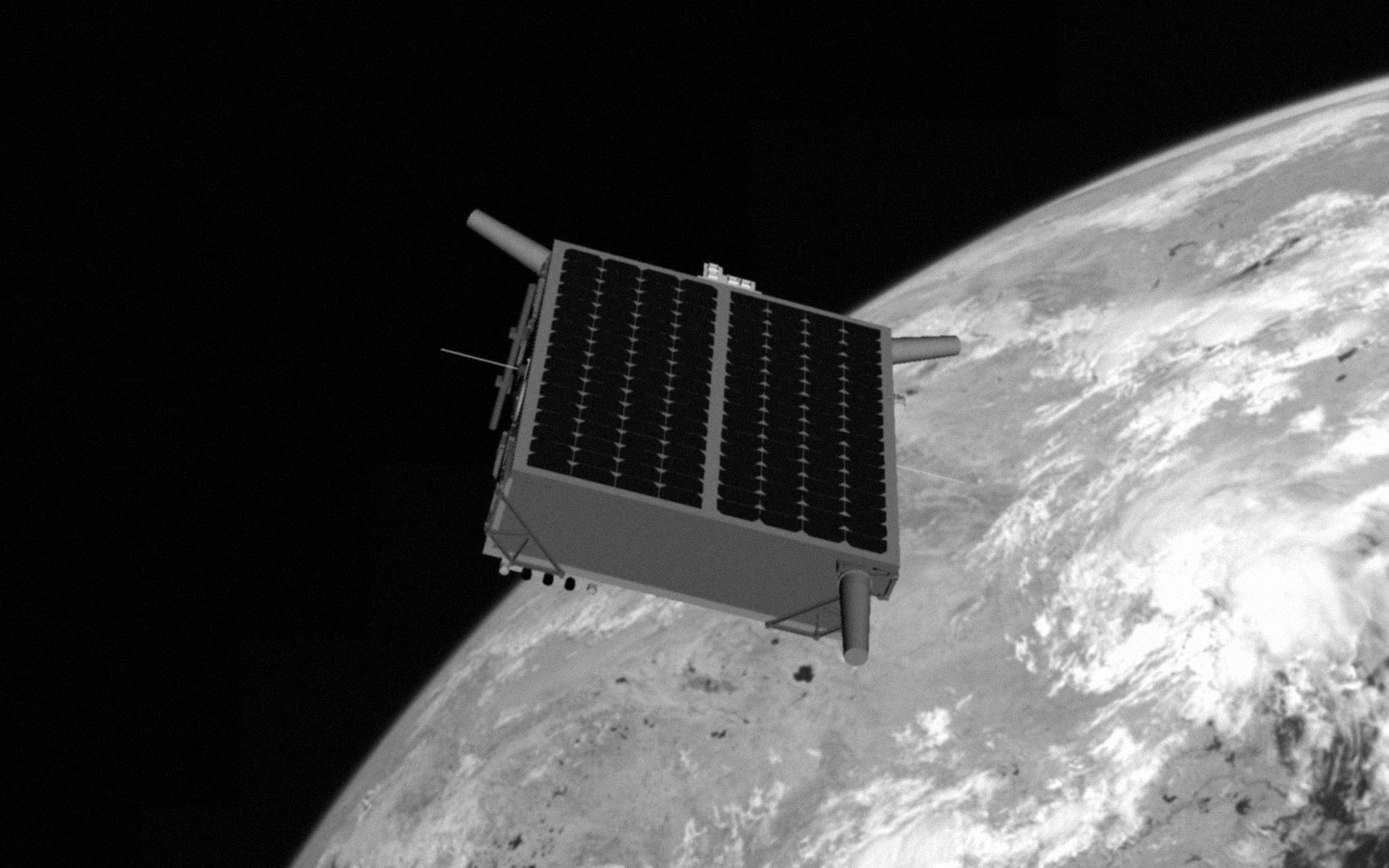} & \includegraphics[width=.14\linewidth]{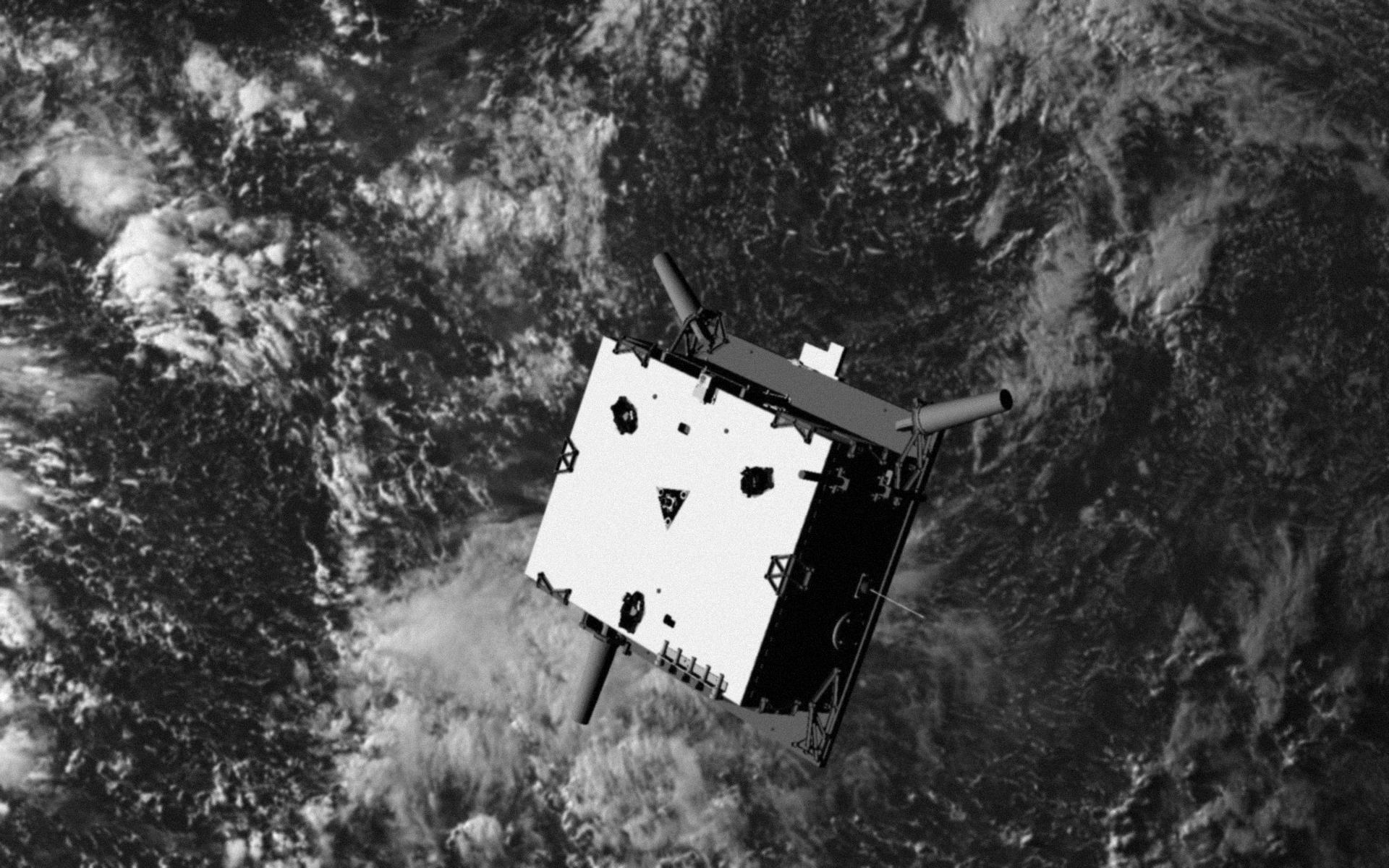} & \includegraphics[width=.14\linewidth]{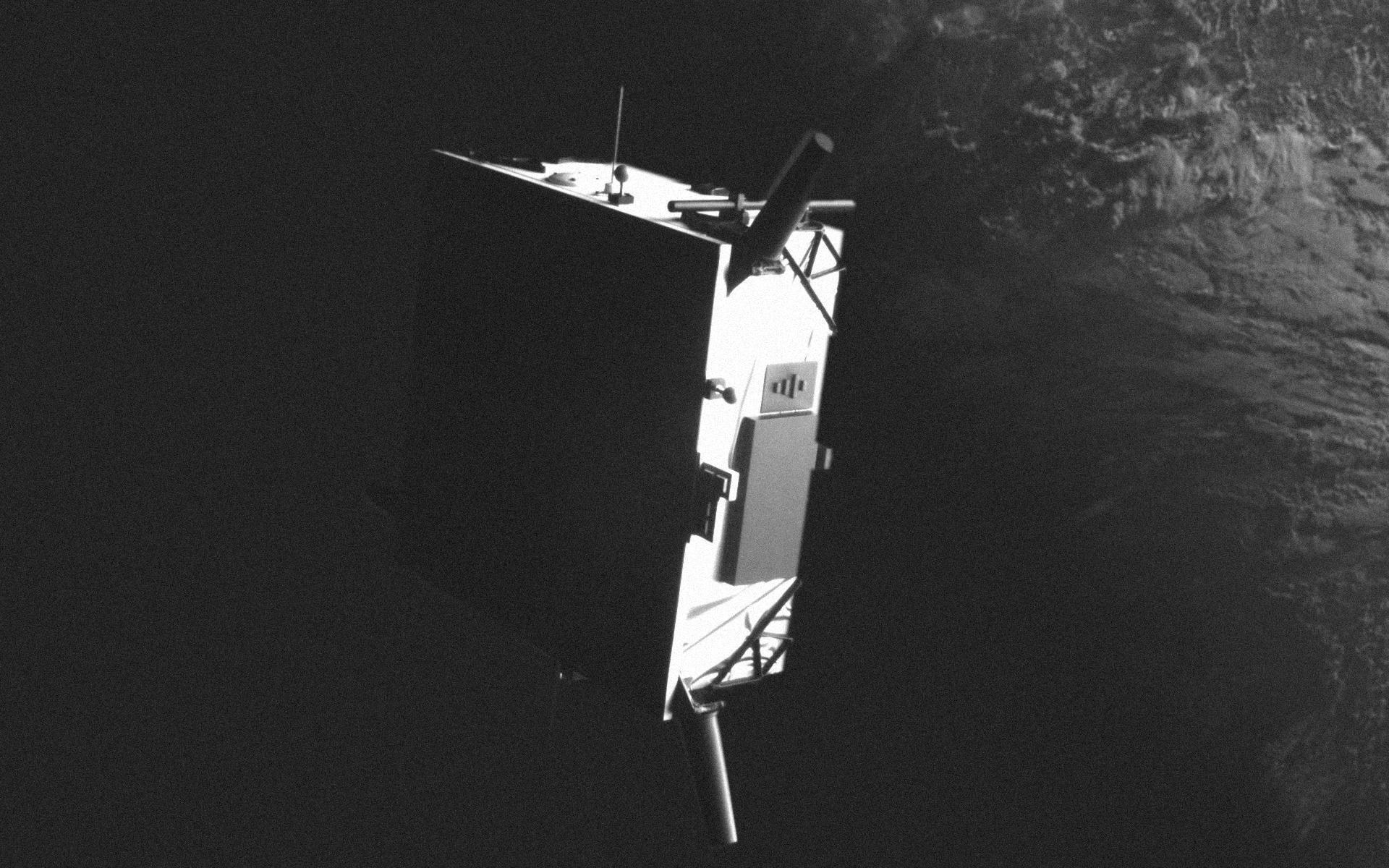} & \includegraphics[width=.14\linewidth]{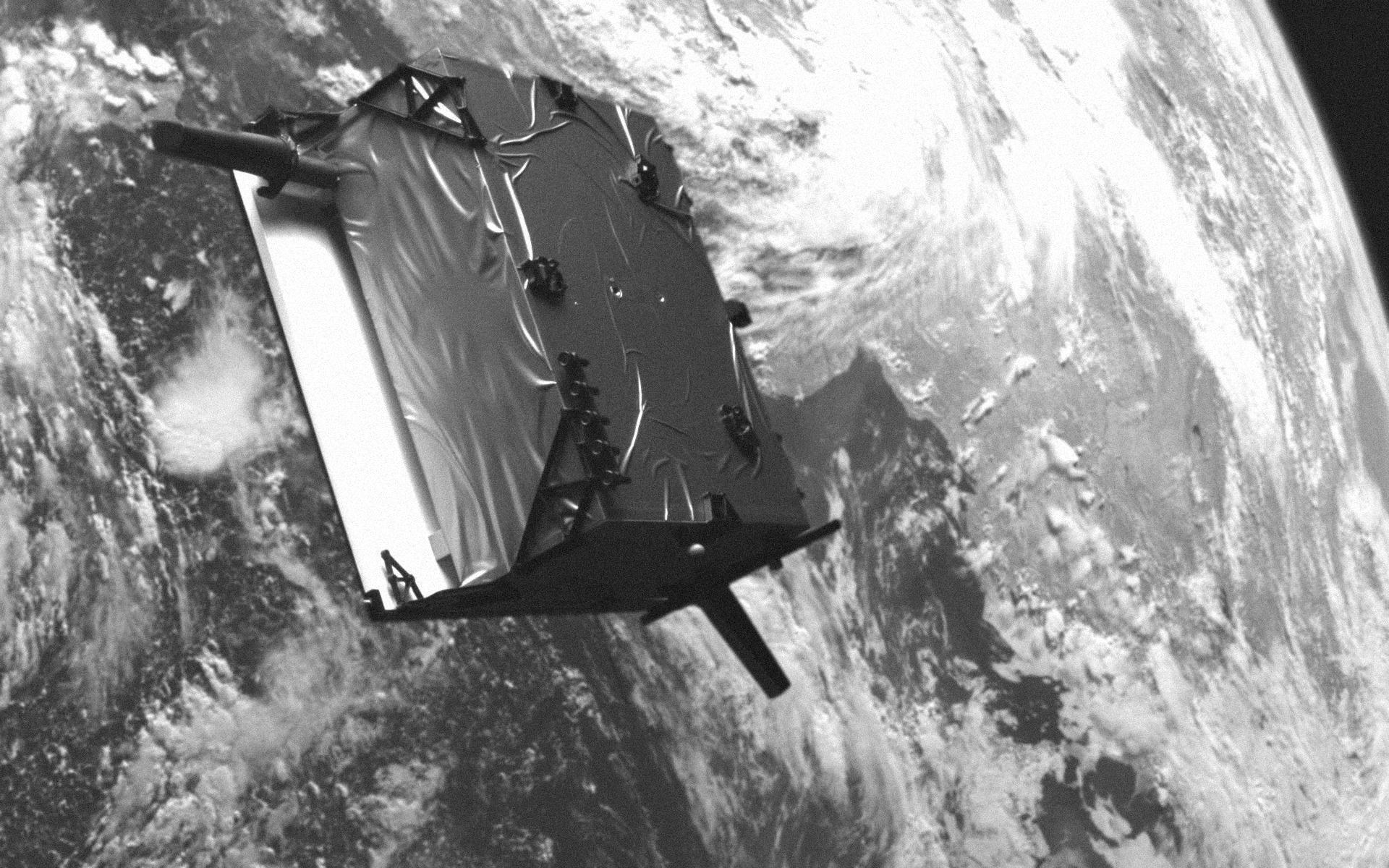}& \includegraphics[width=.14\linewidth]{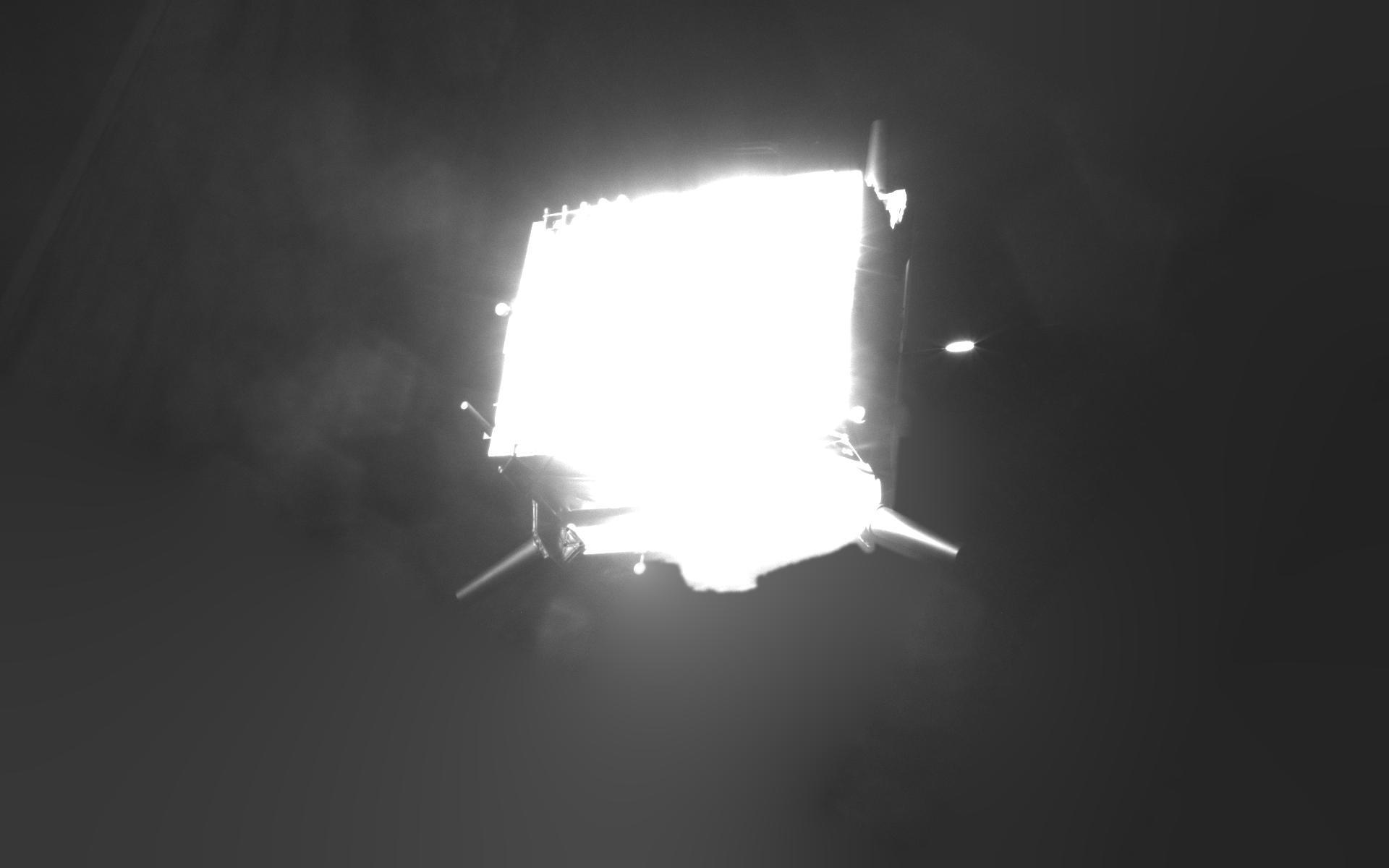}& 
            \includegraphics[width=.14\linewidth]{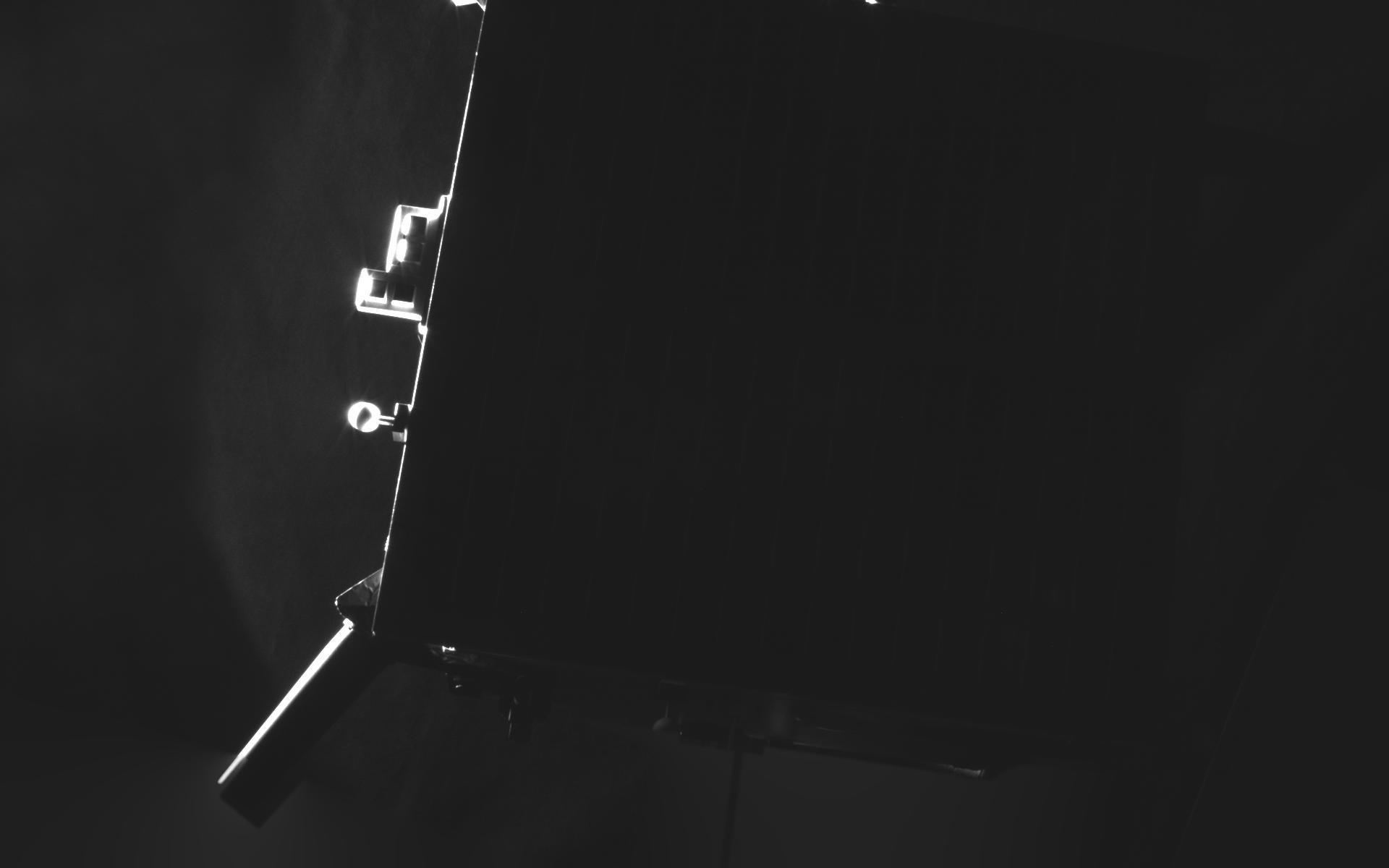}\\   
            \hline
    \end{tabular}    
\end{table*}

\section{Experiments}

    This section assesses the effectiveness of our domain generalization strategy. \Cref{sec_setup} describes the dataset, metrics and network settings used in our experiments. \Cref{sec_exp_analysis} quantitatively evaluates the accuracy of our method and compares it with previous works. Finally, \Cref{sec_exp_ablation} provides ablation studies that demonstrate the impact of the components of our method on its ability to bridge the domain gap.

    \subsection{Experiment setup and evaluation metrics}
    \label{sec_setup}
    
        The following experiments were conducted on the SPEED~\cite{sharma2019spnv1} and SPEED+~\cite{park2022speed+} datasets, used in the Spacecraft Pose Estimation Challenges (SPECs) 2019~\cite{kisantal2020spec2019} and 2021~\cite{park2023spec2021}, respectively, co-organized by the European Agency and Stanford University. SPEED+~\cite{park2022speed+} includes both synthetic and Hardware-In-the-Loop (HIL) images depicting TANGO, the target spacecraft of the PRISMA mission~\cite{gill2007autonomous}, as well as the associated pose labels, while SPEED~\cite{sharma2019spnv1} only contains synthetic images depicting the same target and the associated pose labels. \Cref{tab_datasets} illustrates the three different domains encountered in SPEED+~\cite{park2022speed+}.

        The HIL images of SPEED+~\cite{park2022speed+}, dubbed "\textit{real}" in the rest of this paper, were captured in the TRON facility~\cite{park2021robotic} which mimics the lighting conditions observed in Low Earth Orbit. Different illumination conditions are replicated by \textit{lightbox} and \textit{sunlamp}, the two HIL test set of SPEED+. \textit{Lightbox} is made of 6740 images where the Earth albedo is replicated using light boxes while \textit{sunlamp} is made of 2791 images where the direct illumination of the Sun is replicated using a metal halide lamp. In both domains, the distance from the target to the servicer ranges from 2.5 to 9.5 meters.

        59,960 grayscale images of resolution 1920x1200 were generated through an OpenGL pipeline to form the synthetic set of SPEED+~\cite{park2022speed+}. The images were then processed with Gaussian blur and white Gaussian noise. In half of them, real satellite images of Earth were added in background. The distance from the target to the servicer ranges from 2.2 to 10 meters. SPEED~\cite{sharma2019spnv1} contains 15,000 synthetic images generated under the same conditions as in SPEED+, apart from the distance of the target which ranges from 3m to 50 meters. In the following experiments, to further increase the number of images seen by the network, both synthetic sets are combined and split into training and validation sets ($80\%$-$20\%$). Face segmentation masks are not available in SPEED+. Nevertheless, they are approximated using the keypoint coordinates of the spacecraft body. For each face of each training image, the coordinates of its 4 corners determines a quadrilateral. For each point within that quadrilateral, the point is considered as part of the face mask only if it is not occluded by another face. The purpose of the face segmentation mask being limited to encouraging the network to learn a geometry-aware representation, the supervision of the auxiliary face segmentation task does not require perfect labels.

        In all experiments, the ROI is cropped based on the ground-truth keypoint locations. The KPN and PEM are trained according to the strategy described in \Cref{sec_method} using the Adam~\cite{kingma2014adam} optimizer and cosine annealing~\cite{loshchilov2016cosannealing} with an initial learning rate of $1e^{-3}$, a momentum of $0.9$ and a batch size of 64 images. The KPN is trained with a weight decay of $1e^{-3}$ for 400 epochs while the PEM is trained with a weight decay of $1e^{-4}$ for 50 epochs. Following common practices on SPEED~\cite{chen2019dlr,park2019krn}, the $K=11$ keypoints considered in the pipeline are the 8 corners of the spacecraft body and the top of its three antennas. The KPN input resolution is fixed to 256x256. The PEM embeddings contain 64 features. Its coordinate MLP contains a single hidden layer of 256 neurons. The attention-based encoder is made of six encoder layers, each of them consisting of a self-attention layer and a feed-forward layer of dimension 256.

        Our pipeline is evaluated on the SPEED+ score, $S^{*}_{P}$, introduced in SPEC2021~\cite{park2023spec2021}, that averages the pose scores on the N samples of a test set. For each image in the set, the translation error, $e_{t}$, is computed as the norm of the difference between the predicted and ground-truth positions, $t$ and $\hat{t}$, respectively, \ie,
        \begin{equation}
            e_{t} = \left |\left | t - \hat{t}  \right |\right |,
        \end{equation}
        while the normalized translation error, $\bar{e}_t$, is defined as the ratio between the translation error and the norm of the ground-truth position, \ie,
        \begin{equation}
            \bar{e}_{t} = \frac{e_{t}}{\left |\left |  \hat{t}  \right |\right |}.
        \end{equation}
        The rotation error, $e_{q}$ is the angular error between the predicted and ground-truth quaternions, $q$ and $\hat{q}$, respectively, \ie, 
        \begin{equation}
            e_{q} = 2 \textnormal{ arccos} \left( \left| \hat{q} q^{T}\right| \right).
        \end{equation}
        Since the ground-truth positions and rotations on the HIL domains are not perfectly accurate because of a limited calibration accuracy, the estimates are considered as perfect if they are below a certain threshold, \ie,
        \begin{equation}
          e^{*}_{q} = 
          \begin{cases}
            0, & \text{if } e_{q} < 0.00295 \textnormal{rad}\\
            e_{q}, & \text{otherwise},
          \end{cases} 
        \end{equation}
        and,
        \begin{equation}
          \bar{e}^{*}_{t} = 
          \begin{cases}
            0, & \text{if } \bar{e}_{t} < 2.173 \textnormal{mm/m}\\
            \bar{e}_{t}, & \text{otherwise}
          \end{cases}
        \end{equation}
        Three metrics are used to evaluate the accuracy of the method. The average translation and rotation errors, $E_{T}$ and $E_{R}$, respectively, are computed as the average of the images translation and rotation errors over the test set, \ie,
        \begin{equation}
            E_{T} =  \frac{1}{N} \sum_{i=1}^{N} e^{(i)}_{t},
          \quad \textnormal{and} \quad 
            E_{R} =  \frac{1}{N} \sum_{i=1}^{N} e^{(i)}_{q}.
        \end{equation}
        Finally, the SPEED+ score, $S^{*}_{P}$, is defined as the sum of the normalized translation and rotation errors, averaged over the whole test set, \ie
        \begin{equation}
            S^{*}_{P} =  \frac{1}{N} \sum_{i=1}^{N} \left( \bar{e}^{*(i)}_{t} + e^{*(i)}_{q} \right).
        \end{equation}

\begin{table*}[t]
    \caption{\label{tab_comparison_soa}Overview of the performance metrics averaged on the test datasets for different methods. Our method, which relies on a domain generalization strategy, outperforms the other domain generalization strategies, except EagerNet~\cite{ulmer20236d} which is more complex than the proposed approach. Furthermore, our method is on par with most domain adaptation strategies. Finally, our Lite-HRNet based variant achieves decent performance given its reduced computational load. \vspace{0.2cm}}
    \renewcommand{\arraystretch}{1.0}
    \centering
        \begin{tabular}{|l|c|c|ccc|ccc|}
            \hline
            \rule{0pt}{10pt}
            Method & Domain & Complexity & \multicolumn{3}{c|}{Lightbox} & \multicolumn{3}{c|}{Sunlamp}\\            
            & & [GFlops] & $E_T[m]$ & $E_R[^{\circ}]$ & $S^{*}_{P}[/]$ & $E_T[m]$ & $E_R[^{\circ}]$ & $S^{*}_{P}[/]$\\
            \cline{1-9}  
            \rule{0pt}{10pt}
            \textit{TangoUnchained}~\cite{park2023spec2021} & Adaptation & $\gg$ Ours & 0.11 & 3.19 & 0.07 & 0.09 & 4.30 & 0.09\\
            P{\'e}rez-Villar \etal~\cite{perez2022spacecraft} & Adaptation & $\gg$ Ours & - & 4.59 & 0.10 & - & 2.81 & 0.06 \\
            Wang \etal~\cite{wang2023bridging} & Adaptation & $\gg$ Ours & - & 6.66 & 0.16 & - & 2.73 & 0.06 \\
            SPNv2~\cite{park2023robust} w. ODR & Adaptation & 20.8 & 0.15 & 5.58 & 0.12 & 0.16 & 9.79 & 0.20\\
            \cline{1-9}  
            \rule{0pt}{10pt}
            EagerNet~\cite{ulmer20236d} & Generalization & >19 & 0.09 & 1.75 & 0.04 & 0.013 & 2.66 & 0.06 \\
            SPNv2~\cite{park2023robust} w/o. ODR & Generalization & 20.8 & 0.22 & 7.99 & 0.17 & 0.23 & 10.37 & 0.22\\
            \hline 
            \rule{0pt}{10pt}
            \textbf{Ours} (HRNet) & Generalization & 6.3 & \textbf{0.09} & \textbf{4.32} & \textbf{0.09} & \textbf{0.14} & \textbf{6.94} & \textbf{0.14} \\
            \textbf{Ours} (Lite-HRNet) & Generalization & 1.2 & \textbf{0.14} & \textbf{7.42} & \textbf{0.15} & \textbf{0.20} & \textbf{15.12} & \textbf{0.30} \\
            \hline
        \end{tabular}
\end{table*}

\begin{table*}[b]
    \addtocounter{table}{1} 
    \caption{\label{tab_ablation_multitask} Average performance metrics achieved on \textit{Lightbox} and \textit{Sunlamp} for different multi-task learning strategies, \ie, considering (i) no auxiliary task, (ii) only spacecraft segmentation, or (iii) only faces segmentation, or (iv) both segmentation tasks as auxiliary tasks. The network generalizes better when trained on both auxiliary tasks. \vspace{0.2cm}}
    \centering
    \begin{tabular}{@{\extracolsep{\fill}}|c|ccc|ccc|} 
        \hline
        \rule{0pt}{10pt}
        Aux. Segm.& \multicolumn{3}{c|}{Lightbox} & \multicolumn{3}{c|}{Sunlamp} \\
        Tasks & $E_T[m]$ & $E_R[^{\circ}]$ & $S^{*}_{P}[/]$ & $E_T[m]$ & $E_R[^{\circ}]$ & $S^{*}_{P}[/]$ \\
        \hline  
        \rule{0pt}{10pt}
        No Aux. Task &  0.13 & 7.40 & 0.15 & 0.18 & 12.87 & 0.26 \\ 
        Spacecraft & 0.10 & 5.22 & 0.11 & 0.14 & 9.47 & 0.19 \\
        Faces & 0.11 & 4.77 & 0.10 & 0.15 & 9.67 & 0.19 \\
        \hline  
        \rule{0pt}{10pt}
        Both & \textbf{0.09} & \textbf{4.32} & \textbf{0.09} & \textbf{0.14} & \textbf{6.94} & \textbf{0.14} \\
        \hline  
    \end{tabular}
\end{table*} 

    \subsection{Evaluation}
    \label{sec_exp_analysis}
    
        As pointed out in \Cref{tab_comparison_soa}, our method successfully bridges the domain gap. It achieves average errors of 9.3cm and \ang{4.3} on \textit{Lightbox} and 14cm and \ang{6.9} on \textit{Sunlamp}. On \textit{Lightbox}, the median errors are of 6cm and \ang{1.6} while those median errors are of 9cm and \ang{2.5} on \textit{Sunlamp}. On the SPEED+ synthetic validation set, the median errors are of 3 cm and \ang{1}. Since those median errors are not significantly better than the ones obtained on \textit{Lightbox} and \textit{Sunlamp}, we conclude that the proposed approach successfully bridges the domain gap.

        \Cref{tab_comparison_soa} also summarizes the accuracy obtained by State-of-The-Art methods. Our method outperforms all those which follow a domain generalization strategy, \ie, which do not rely on the knowledge of the test set at training, except for the EagerNet method~\cite{ulmer20236d}, which is computationally more intensive. Furthermore, our method is on par with most domain adaptation approaches, \ie, which exploits the test domains during the network training. This highlights that a proper learning strategy can bridge the domain gap without using images from the target domain.

        \Cref{tab_comparison_soa} also highlights the accuracy of our method used with a Lite-HRNet~\cite{yu2021litehrnet} backbone. Even, if this network is not as accurate as its HRNet-based~\cite{sun2019hrnet} equivalent, it is less computationally intensive, \ie, the number of operations is divided by 5, while only doubling the errors. Given the low computing capabilities offered by space-grade hardware, it therefore presents an interest for future missions. This further  demonstrates that large neural networks are not required to successfully bridge the domain gap, contrary to what is commonly accepted~\cite{cassinis2022monocular}.

    \subsection{Ablation Studies}
    \label{sec_exp_ablation}     
        This section aims at investigating the role of key components of our approach in its final accuracy. Sections \ref{sec_exp_ablation_Equalization}, \ref{sec_exp_multitask} and \ref{sec_exp_domainRand} analyze the impact of, respectively, the histogram equalization, the multi-task learning and the domain randomization, on the generalization abilities of our method. Finally, \Cref{sec_exp_ablation_Architecture} focuses on the PEM architecture. 

        \subsubsection{Histogram Equalization}
        \label{sec_exp_ablation_Equalization}
            Performing histogram equalization on the KPN input aims at reducing the distance between the test and training domains by enforcing the histograms of all images, \ie, from both train and test domains, to follow an uniform distribution. As a result, the parts of the images that are over-exposed are shadowed while the ones that are under-exposed appear brighter. This decreases the gap between the domains and therefore improves the generalization abilities of the network. As pointed out in \Cref{tab_ablation_equalization}, this equalization decreases the errors by $25\%$ on both sets.

\begin{table}[h]
    \setcounter{table}{2} 
    \caption{\label{tab_ablation_equalization} Average performance metrics achieved by our method on \textit{Lightbox} and \textit{Sunlamp} with or without performing histogram equalization on the KPN input. Histogram Equalization improves the generalization abilities of our method.  \vspace{0.2cm}}
    \centering
    \begin{tabular}{@{\extracolsep{\fill}}|c|ccc|ccc|} 
        \hline
        \rule{0pt}{10pt}
         & \multicolumn{3}{c|}{Lightbox} & \multicolumn{3}{c|}{Sunlamp} \\
        \hline  
        \rule{0pt}{10pt}
        Equal. & $E_T[m]$ & $E_R[^{\circ}]$ & $S^{*}_{P}[/]$ & $E_T[m]$ & $E_R[^{\circ}]$ & $S^{*}_{P}[/]$ \\
        \hline  
        \rule{0pt}{10pt}
        \ding{53} & 0.11 & 6.20 & 0.13 & 0.14 & 8.55 & 0.17 \\
        \hline  
        \rule{0pt}{10pt}
        \Checkmark & \textbf{0.09} & \textbf{4.32} & \textbf{0.09} & \textbf{0.14} & \textbf{6.94} & \textbf{0.14} \\
        \hline
    \end{tabular}
\end{table} 

\begin{table*}[h]
    \addtocounter{table}{1} 
    \caption{\label{tab_ablation_data_augmentation} Average performance metrics achieved on \textit{Lightbox} and \textit{Sunlamp} for data different data augmentation strategies. While our network trained without Data Augmentation does not generalize at all, the same network trained with our Domain Randomization technique achieves a state-of-the-art accuracy. This demonstrates the interest of aggressive data augmentation policies for Domain Generalization. The two augmentations that contribute the most to the generalization capabilities are the texture and exposure ones. \vspace{0.2cm}}
    \centering
    \begin{tabular}{@{\extracolsep{\fill}}|ccccc|ccc|ccc|} 
        \hline
        \rule{0pt}{10pt}
        & \multicolumn{3}{c}{Data Augmentation Techniques} & & \multicolumn{3}{c|}{Lightbox} & \multicolumn{3}{c|}{Sunlamp} \\
        \hline  
        \rule{0pt}{10pt}
        Brightness & Hide\&Seek & Exposure & Texture & Noise & $E_T[m]$ & $E_R[^{\circ}]$ & $S^{*}_{P}[/]$ & $E_T[m]$ & $E_R[^{\circ}]$ & $S^{*}_{P}[/]$ \\
        \hline  
        \rule{0pt}{10pt}
         & & & & & 0.51 & 40.71 & 0.79 & 1.64 & 106.33 & 2.14 \\ 
        \hline  
        \Checkmark & & & & & 0.24 & 18.32 & 0.36 & 0.90 & 79.91 & 1.56 \\ 
         & \Checkmark & & & & 0.18 & 14.24 & 0.28 & 0.632 & 74.22 & 1.40 \\ 
         & & \Checkmark & & & 0.53 & 34.47 & 0.69 & 2.318 & 35.5 & 1.82 \\ 
         & & & \Checkmark & & 0.18 & 9.29 & 0.19 & 0.373 & 26.89 & 0.53\\ 
         & & & & \Checkmark & 0.29 & 18.79 & 0.38 & 1.544 & 87.40 & 1.78 \\ 
        \hline  
        & \Checkmark & \Checkmark & \Checkmark & \Checkmark & 0.11 & 5.55 & 0.11 & 0.15 & 9.18 & 0.19 \\ 
        \Checkmark & & \Checkmark & \Checkmark & \Checkmark & 0.10 & \textbf{4.31} & 0.09 & 0.14 & 7.83 & 0.16 \\ 
        \Checkmark & \Checkmark & & \Checkmark & \Checkmark & 0.10 & 4.75 & 0.10 & 0.17 & 12.03 & 0.24 \\ 
        \Checkmark & \Checkmark & \Checkmark & & \Checkmark & 0.10 & 4.72 & 0.10 & 0.17 & 12.37 & 0.24 \\ 
        \Checkmark & \Checkmark & \Checkmark & \Checkmark & & 0.12 & 6.25 & 0.13 & \textbf{0.14} & 7.39 & 0.15 \\ 
        \hline  
        \Checkmark & \Checkmark & \Checkmark & \Checkmark & \Checkmark & \textbf{0.09} & 4.32 & \textbf{0.09} & \textbf{0.14} & \textbf{6.94} & \textbf{0.14} \\
        \hline  
    \end{tabular}
\end{table*} 

        \subsubsection{Multi-Task Learning}
        \label{sec_exp_multitask}

        As stated in \Cref{sec_multi_task}, our domain generalization strategy relies on multi-task learning to force our network to learn more generic features through auxiliary tasks. At inference, the auxiliary tasks are not inferred, so that the multi-task learning strategy does not increase the computational complexity of the network at inference. Our method relies on both spacecraft segmentation and faces segmentation as auxiliary tasks. As pointed out in \Cref{tab_ablation_multitask}, adding an auxiliary task decreases the errors by $30\%$ on \textit{Lightbox} and $25\%$ on \textit{Sunlamp}, compared to a network trained on the heatmap-based keypoint regression task only. Furthermore, if the network is trained on both segmentation tasks, the errors are decreased by $40\%$ on \textit{Lightbox} and $45\%$ on \textit{Sunlamp}.

        \subsubsection{Domain Randomization}
        \label{sec_exp_domainRand}
            As explained in \Cref{sec_domain_randomization}, our domain generalization strategy relies on domain randomization to enlarge the training set distribution in order to force the network to learn domain-invariant features through aggressive data augmentation techniques. Those data augmentations consist of Gaussian Noise, Brightness \& Contrast, Hide\&Seek~\cite{singh2018hide}, texture and exposure~\cite{sakkos2019illumination} augmentations, resulting in distorted training images. \Cref{fig_Domain_Randomization_examples} provides some examples of training images augmented through this domain randomization strategy. 

            \Cref{tab_ablation_data_augmentation} presents the average errors achieved by our method trained using different data augmentation strategies. With no data augmentation, the network completely overfits on the synthetic train set and provides average errors of 0.51m and \ang{40.7} on \textit{Lightbox} and 1.6m and \ang{106} on \textit{Sunlamp}. The testing images are so different from the images seen during training that the network has not learnt any robust features and does not generalize at all. \Cref{tab_ablation_data_augmentation} also provides the average errors achieved by the network when it is trained with a single data augmentation randomly applied to 50\% of the images. While the usual augmentations, \ie, brightness \& contrast, hide\&seek and Gaussian noise, reduce the average errors by 55\% on \textit{Lightbox} and 25\% on \textit{Sunlamp}, our texture-based augmentation achieves a reduction of 75\% of the average pose score on both test sets. Interestingly, the exposure augmentation brings little gain, \ie, 13\% on \textit{Lightbox} while it decreases by 75\% the pose score on \textit{Sunlamp}. Combining these augmentations through a RandAugment~\cite{cubuk2020randaugment} policy significantly reduces the pose score, \ie, by 89\% on \textit{Lightbox} and 93\% on \textit{Sunlamp}. 

            Finally, \Cref{tab_ablation_data_augmentation} also shows the average errors achieved when the network is trained using RandAugment~\cite{cubuk2020randaugment} with all data augmentation policies but one. The augmentations that contribute the most to the generalization abilities of the network are the texture and exposure ones. Without either of these techniques, the average errors on \textit{Sunlamp} increase by 65\%. This can be explained from the fact that the exposure augmentation produces artifacts that are close to the ones observed in \textit{Sunlamp} while the texture augmentation is well suited for this domain randomization task because it preserves the semantics of the image while randomizing its style.
  
\begin{table*}[t]
    \caption{\label{tab_ablation_architecture} Average performance metrics achieved by our method on \textit{Lightbox} and \textit{Sunlamp}, considering simplified Pose Estimation Models. (a) consists in a simple MLP applied on all keypoint coordinates. (b) adds to (a) a mapping of the keypoint coordinates to an embedding in a higher dimensional space. (c) processes those embeddings through an attention-based encoder before passing them to the MLP. (d) adds positional embeddings at the input of the encoder. Finally, the benefits of the continuous 6D rotation representation over the quaternion representation are evaluated. \vspace{0.2cm}}
    \centering
    \begin{tabular}{@{\extracolsep{\fill}}|l|c|ccc|ccc|} 
        \hline
        \rule{0pt}{10pt}
        PEM Arch. & Rotation  & \multicolumn{3}{c|}{Lightbox} & \multicolumn{3}{c|}{Sunlamp} \\
         & format & $E_T[m]$ & $E_R[^{\circ}]$ & $S^{*}_{P}[/]$ & $E_T[m]$ & $E_R[^{\circ}]$ & $S^{*}_{P}[/]$ \\
        \hline  
        \rule{0pt}{10pt}
        (a): MLP & 6D & 0.12 & 4.60 & 0.10 & 0.16 & 7.38 & 0.16 \\
        (b): (a)+EMB & 6D & 0.11 & 4.55 & 0.10 & 0.16 & 7.32 & 0.16 \\
        (c): (b)+ENC & 6D & 0.10 & 4.37 & 0.09 & 0.14 & 7.08 & 0.15 \\
        (d): (c)+PE & 6D & \textbf{0.09} & \textbf{4.32} & \textbf{0.09} & \textbf{0.14} & \textbf{6.94} & \textbf{0.14} \\
        \hline\rule{0pt}{10pt}
        (d) & 4D & 0.11 & 7.65 & 0.15 & 0.14 & 10.10 & 0.20\\
        \hline
    \end{tabular}
\end{table*} 

        \subsubsection{PEM Architecture}
        \label{sec_exp_ablation_Architecture}
            Both HRNet~\cite{sun2019hrnet} and Lite-HRNet~\cite{yu2021litehrnet} have been used in spacecraft pose estimation pipelines~\cite{chen2019dlr,carcagni2022lightweight}. The novelty of our approach comes from the attention-based Pose Estimation Model used to predict the 6D pose from a set of predicted keypoint coordinates using a continuous 6D rotation representation~\cite{zhou2019continuity}. This section evaluates the impact of some of its design choices on its accuracy.

            \Cref{tab_ablation_architecture} presents the average errors achieved by our method considering several variations from the proposed PEM architecture on two test sets, \ie, \textit{Lightbox} and \textit{Sunlamp}. It shows that even if directly predicting the pose from the keypoints through a MLP works, expanding the dimension of the keypoints before processing them through this MLP improves the performances on all sets. In addition, pre-processing the keypoints through an attention-based encoder further improves the accuracy. Finally, adding a positional embedding to the keypoint embeddings provides additional gains. \Cref{tab_ablation_architecture} also shows that the use of the continuous 6D representation for regressing rotations significantly increases the accuracy of the network. 
        
\section{Conclusion}
    This paper addressed the domain gap problem of the spacecraft pose estimation task. Our solution follows a usual architecture in spacecraft pose estimation, \ie, heatmap-based keypoint regression followed by pose estimation. This two-step approach concentrates the image processing generalization issue on the keypoint regression part, and leverages on the robustness of the pose estimation module to deal with erroneous keypoint predictions. In our work, the pose estimation exploits an attention-based encoder to achieve robust pose prediction, in the form of a continuous 6D rotation and normalized 3D position, from the keypoint coordinates.

    We introduced an efficient learning strategy that relies on histogram equalization, multi-task learning and domain randomization \ie, aggressive data augmentations.
    Unlike previous works that followed a domain adaptation strategy, \ie, aimed at reducing the gap by exploiting the target domain during the network training, our domain generalization strategy enables the training of our network using only synthetic images, and without increasing the inference computational cost. We successfully validated our method on the widespread SPEED+ dataset. The roles of the different components of our approach were evaluated through extensive ablation studies.

\section*{Funding Sources}
The research was funded by Aerospacelab and the Walloon Region through the Win4Doc program. C. De Vleeschouwer is a Research Director of the Fonds de la Recherche Scientifique – FNRS.

\section*{Acknowledgments}
Computational resources have been provided by the Consortium des Équipements de Calcul Intensif (CÉCI), funded by the Fonds de la Recherche Scientifique de Belgique (F.R.S.-FNRS) under Grant No. 2.5020.11 and by the Walloon Region.

\normalsize
\bibliography{references}


\end{document}